%% file: googledeepmind-test.tex
\documentclass[11pt, letterpaper, logo]{googledeepmind}

\PassOptionsToPackage{comma,numbers,sort,compress}{natbib}

\usepackage[comma,numbers,sort,compress]{natbib}

\usepackage{hyperref}[citecolor=magenta]
\usepackage{fdsymbol}

\hypersetup{
    colorlinks = true,
    citecolor = {magenta},
}

\pdftrailerid{redacted}

\makeatletter
\renewcommand\bibentry[1]{\nocite{#1}{\frenchspacing\@nameuse{BR@r@#1\@extra@b@citeb}}}
\makeatother

\usepackage{kantlipsum, lipsum}
\usepackage{dsfont}
\usepackage{gdm-colors}
\usepackage{subfigure}
\usepackage{bbm}
\usepackage{wrapfig}

\usepackage[most,skins,theorems]{tcolorbox}

\tcbset{
  aibox/.style={
    width=\linewidth,
    top=8pt,
    bottom=4pt,
    colback=blue!6!white,
    colframe=black,
    colbacktitle=black,
    enhanced,
    center,
    attach boxed title to top left={yshift=-0.1in,xshift=0.15in},
    boxed title style={boxrule=0pt,colframe=white,},
  }
}
\newtcolorbox{AIbox}[2][]{aibox,title=#2,#1}
\definecolor{lightblue}{rgb}{0.22,0.45,0.70}

\graphicspath{{figures/}}

\title{Scaling LLM Test-Time Compute Optimally can be More Effective than Scaling Model Parameters}

\correspondingauthor{csnell22@berkeley.edu}

\author[$\vardiamondsuit$, 1]{Charlie Snell}
\author[2]{Jaehoon Lee}
\author[$\clubsuit$, 2]{Kelvin Xu}
\author[$\clubsuit$, 2]{Aviral Kumar}

\affil[$\clubsuit$]{Equal advising}
\affil[1]{UC Berkeley}
\affil[2]{Google DeepMind}
\affil[$\vardiamondsuit$]{Work done during an internship at Google DeepMind}
\vspace{-0.5cm}

\begin{abstract}
\vspace{-0.4cm}
Enabling LLMs to improve their outputs by using more test-time computation is a critical step towards building generally self-improving agents that can operate on open-ended natural language. In this paper, we study the scaling of inference-time computation in LLMs, with a focus on answering the question: \emph{if an LLM is allowed to use a fixed but non-trivial amount of inference-time compute, how much can it improve its performance on a challenging prompt?} Answering this question has implications not only on the achievable performance of LLMs, but also on the future of LLM pretraining and how one should tradeoff inference-time and pre-training compute. Despite its importance, little research attempted to understand the scaling behaviors of various test-time inference methods. Moreover, current work largely provides negative results for a number of these strategies. In this work, we analyze two primary mechanisms to scale test-time computation: \textbf{(1)} searching against dense, process-based verifier reward models; and \textbf{(2)} updating the model's distribution over a response adaptively, given the prompt at test time. We find that in both cases, the effectiveness of different approaches to scaling test-time compute critically varies depending on the difficulty of the prompt. This observation motivates applying a ``compute-optimal'' scaling strategy, which acts to most effectively allocate test-time compute adaptively per prompt. Using this compute-optimal strategy, we can improve the efficiency of test-time compute scaling by more than \textbf{4$\times$} compared to a best-of-N baseline. Additionally, in a FLOPs-matched evaluation, we find that on problems where a smaller base model attains somewhat non-trivial success rates, test-time compute  can be used to outperform a \textbf{14$\times$} larger model.
\end{abstract}

\begin{document}

\maketitle

\input{template_content}

\bibliographystyle{abbrvnat}
\nobibliography*
\bibliography{template_refs}

\newpage 

\appendix 
\part*{Appendices}

\vspace{-0.2cm}
\section{Related Work}
\vspace{-0.2cm}

\textbf{Language model reasoning.} Language model performance on challenging mathematical reasoning tasks has rapidly improved in recent years~\citep{lewkowycz2022solving,geminiteam2024gemini,openai2024gpt4,shao2024deepseekmath,lightman2023lets}.
These improvements can be attributed to four primary factors: \textbf{1)} running continued pretraining on large corpora of math focused data~\citep{lewkowycz2022solving,geminiteam2024gemini,shao2024deepseekmath,lightman2023lets}; \textbf{2)} improving the LLM proposal distribution by either applying targeted optimization on specific reasoning tasks by finetuning with RL~\citep{singh2024human,zelikman2022star,shao2024deepseekmath,yuan2023scaling} enabling models to critique and revise their answers iteratively~\citep{bai2022constitutional,madaan2023selfrefine,du2023improving,saunders2022selfcritiquing}; \textbf{3)} enabling LLMs to benefit from additional test-time computation by finetuning verifiers~\citep{lightman2023lets,cobbe2021training,uesato2022solving,wang2023mathshepherd,yao2023tree,feng2024alphazerolike,chen2024alphamath,tian2024selfimprovement}.
Our work builds on these second and third lines of research by analyzing the extent to which test-time compute scaling can be improved by 1) refining an LLM's proposal distribution and 2) conducting search against verifiers.

\textbf{Analyzing test-time compute scaling.} The tradeoff between train-time and test-time compute using Monte-Carlo tree search applied to the board game Hex was previously studied by~\citet{jones2021scalingscalinglawsboard}. We instead focus our analysis on full-scale language model math reasoning problems. A survey work by~\citet{epoch2023tradingoffcomputeintrainingandinference} analyzed the tradeoff between training and inference across a number of domains. However, much of their language-model analysis focused on test-time compute scaling in settings where the ground-truth answer is known. In contrast, our analysis focuses on the setting when the ground-truth answer is not known. Additionally, a number of works in the RL literature have proposed methods, such as MCTS~\citep{kocsis2006bandit}, which aim to navigate the tradeoff between test-time and training-time compute so as to enable a form of iterative self-play. The findings in our work can be used to help develop similar algorithms that can operate on open-ended natural language.

\textbf{Augmenting LLMs with test-time compute.} Beyond verifiers and revisions, a number of additional works have proposed alternative methods for enabling LMs to use test-time compute for reasoning. Namely,~\citet{wang2024hypothesissearchinductivereasoning} conducts a hierarchical hypothesis search to enable inductive reasoning capabilities. A number of related works have proposed augmenting language models with tools at test-time, which can greatly improve their performance on downstream tasks~\citep{gao2023palprogramaidedlanguagemodels,qin2023toolllmfacilitatinglargelanguage,qu2024toollearninglargelanguage}.
Finally, several works have proposed methods for learning thought tokens in an unsupervised manner~\citep{zelikman2024quietstarlanguagemodelsteach,goyal2024thinkspeaktraininglanguage}, enabling models to more effectively utilize the additional test-time compute that comes with sampling longer sequences. While we focus our analysis on two primary mechanisms by which test-time compute can be scaled in this work (e.g. verifiers and revisions), many of the methods by which we conduct our analysis (e.g. compute optimal scaling according to question difficulty) could, in principle, also be applied to any of these other methods of scaling test-time compute, and we believe that this is an interesting direction for future research.

\section{Additional Revision Results}
\label{app:additional_revision}

\begin{figure}
    \centering
    \includegraphics[width=0.99\textwidth]{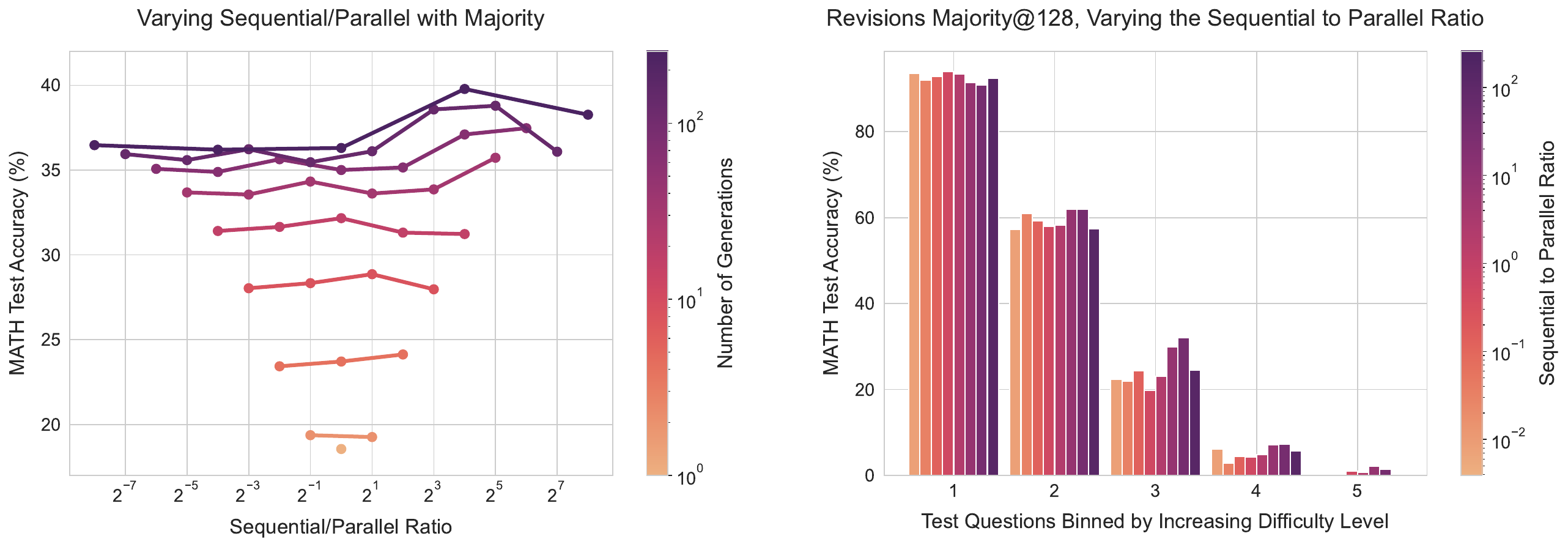}
    \caption{Varying the ratio of generation budget allocated to sequential verses parallel samples, using majority to select the answer, rather than the verifier. \textbf{Left:} Each line represents a fixed generation budget as the ratio is changed. We see that similar to the verifier case, in the majority case, there exists an ideal ratio of  sequential to parallel test-time compute at a given budget. \textbf{Right:} Analyzing performance across difficulty bins, we see that the easier questions are mostly invariant the ratio of sequential to parallel, whereas on the harder questions there is an ideal ratio of sequential to parallel test-time compute.}
    \label{fig:revision_maj_additional}
\end{figure}

We plot additional results for majority selection using out PaLM 2-S* revision model in Figure~\ref{fig:revision_maj_additional}. With majority selection, we see largely similar trends to those found in Figure~\ref{fig:iso_revisions} for verifier selection.

\section{Unsupervised Difficulty Bins}
\label{app:unsupervised_difficulty}

We compute difficulty bins without oracle ground-truth correctness information by averaging the PRM final-answer score over 2048 samples on each question, so as to obtain a value estimate corresponding to the question. We then bin the value for each question in the test-set into five quintiles (using the same procedure as the oracle difficulty bins). We refer to this as ``predicted difficulty'' rather than ``oracle difficulty''. Technically this procedure is extremely costly because it requires generating many samples. While we do not account for this cost in our analysis, in a practical production setting, this cost would be problematic. A more efficient approach would be to finetune a model to predict correctness directly, given the question. We do not explore this in our work, but leave such exploration of cheaper methods of estimating difficulty to future work.

\begin{figure}
    \centering
    \subfigure{
        \includegraphics[width=0.48\textwidth]{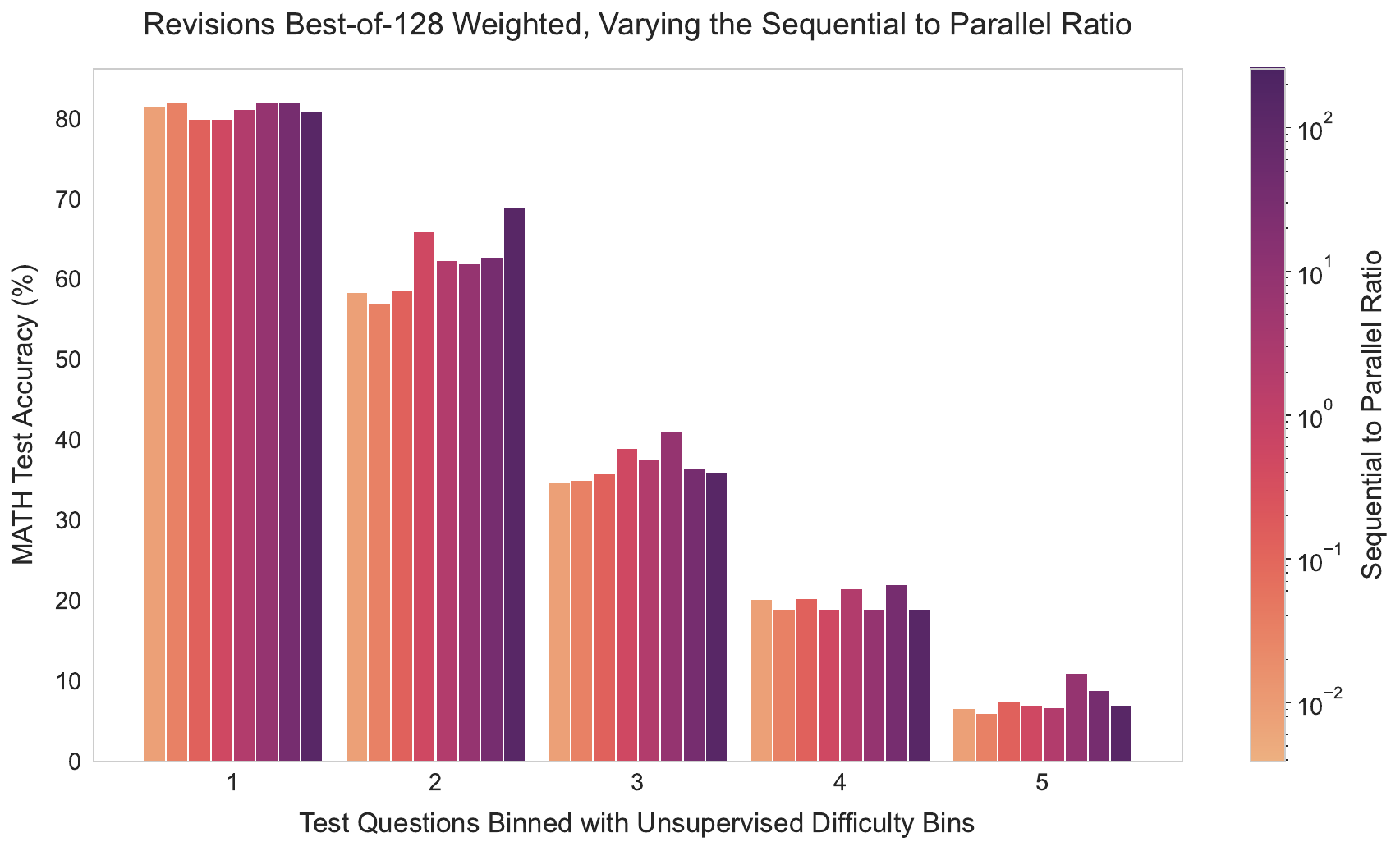}
        \label{fig:bon_unsupervised_difficulty_output}
    }\hfill
    \subfigure{
        \includegraphics[width=0.48\textwidth]{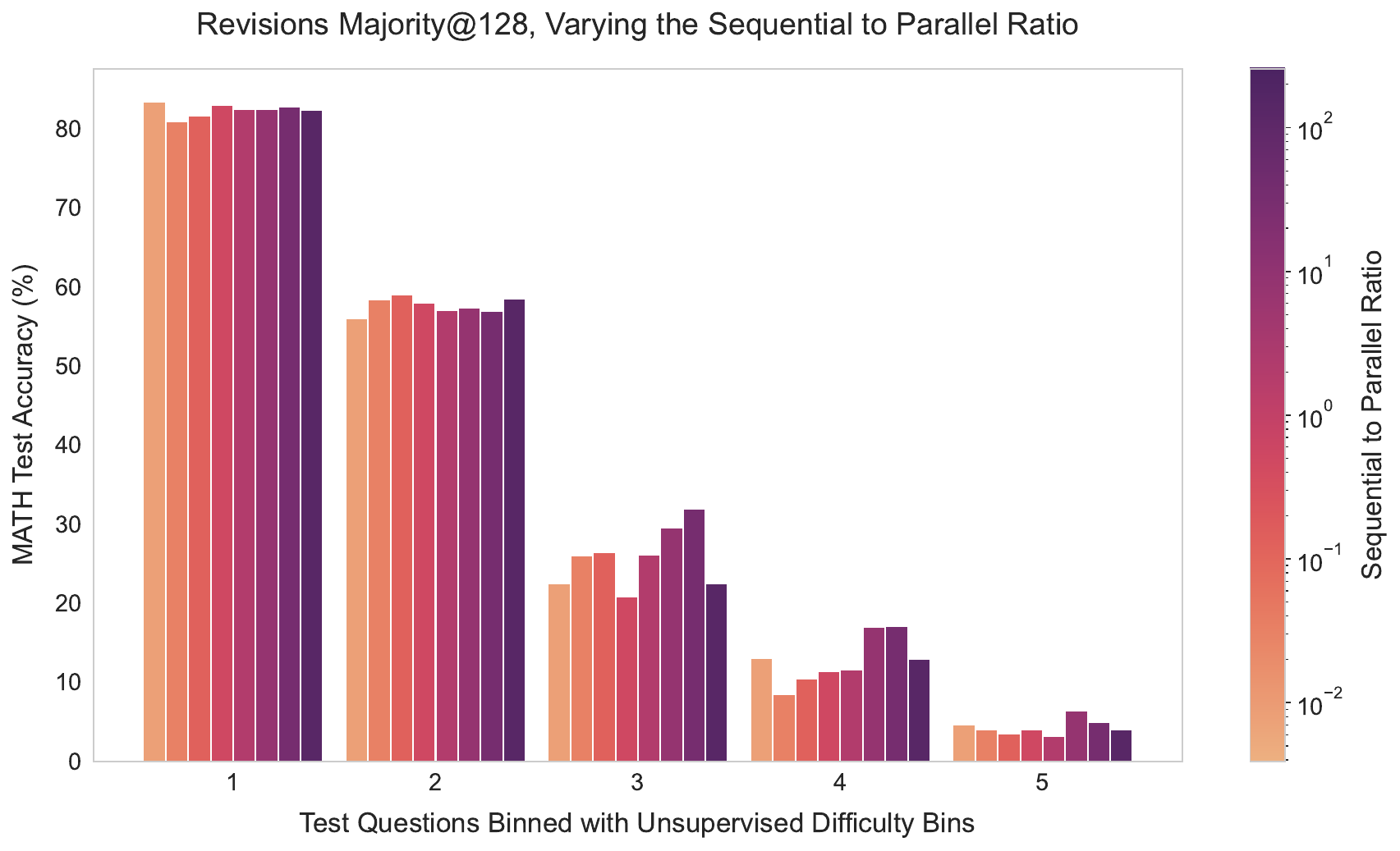}
        \label{fig:maj_unsupervised_difficulty_output}
    }
    \caption{Using our PaLM 2-S* PRM to compute difficulty bins without ground truth correctness information for revisions. On the left we plot verifier selection and on the right we plot majority selectionl We see largely similar performance trends with these bins as we do with the ground truth ones in Figures~\ref{fig:iso_revisions} and~\ref{fig:revision_maj_additional}.}
    \label{fig:revisions_unsupervised_difficulty}
\end{figure}

\begin{figure}
    \centering
    \includegraphics[width=0.6\textwidth]{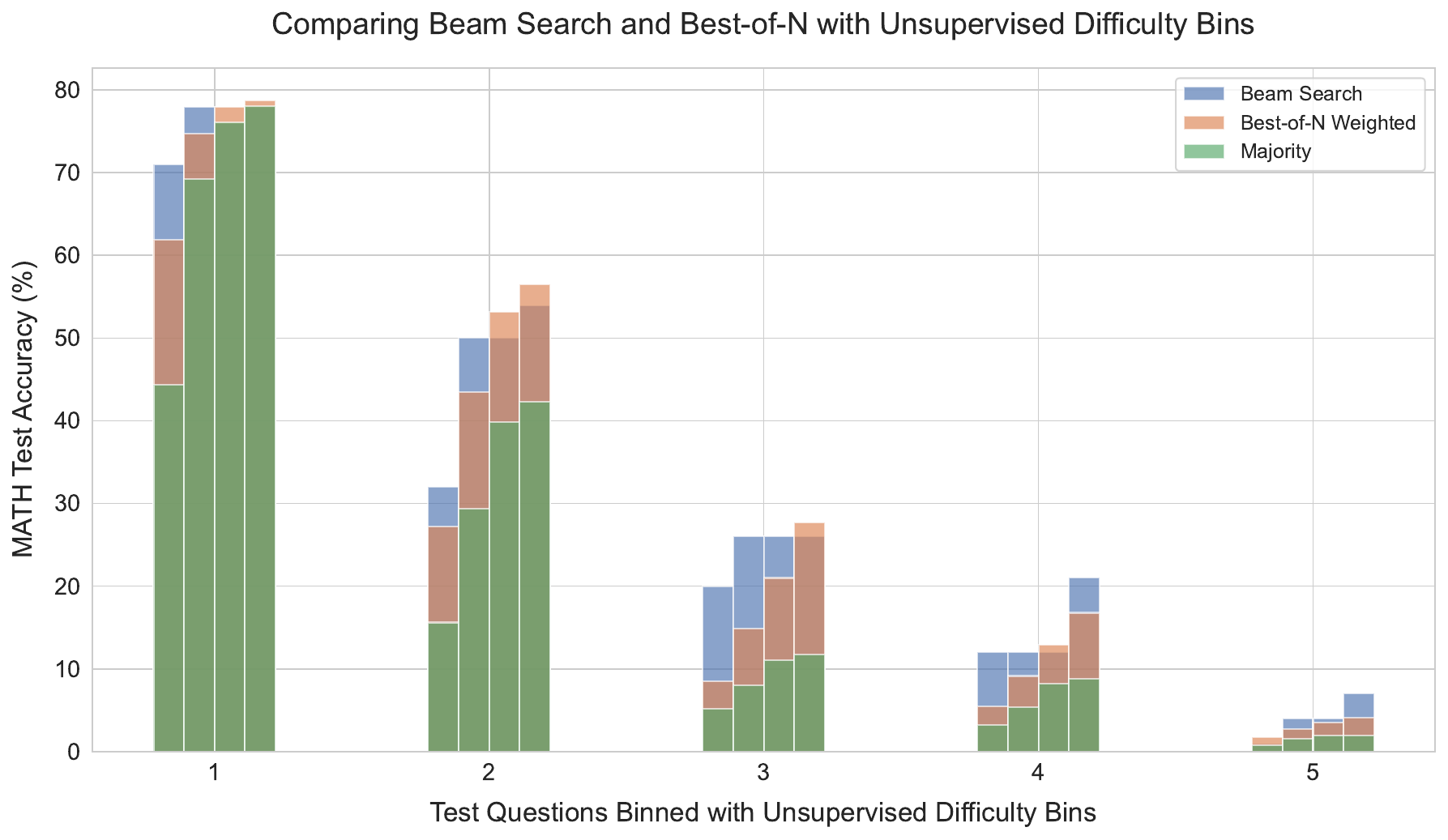}
    \caption{Using our PaLM 2-S* PRM to compute difficulty bins without ground truth correctness information for PRM search. We see largely similar performance trends with these bins as we do with the ground truth ones in Figure~\ref{fig:comparing_search_and_beam_difficulty}.}
    \label{fig:search_unsupervised_difficulty}
\end{figure}

In Figure~\ref{fig:search_unsupervised_difficulty} we plot PRM-search results using our difficulty bins, and in Figure~\ref{fig:revisions_unsupervised_difficulty} we plot the corresponding revision results. We see that in both settings these predicted bins demonstrate similar trends to the oracle bins.

\section{PRM Training Details}
\label{app:prm_training}

We finetune our PRM as a binary classifier, where the model predicts a value between 0 and 1 at each step in the solution. We train the model with soft values obtained from the monte-carlo rollouts, using a binary cross entropy loss function (e.g. $-(ylog(\hat{y})+(1-y)log(1-\hat{y}))$ where $y$ corresponds to the soft ground-truth value and $\hat{y}$ the model's predicted value). We finetune the model base model using the AdamW optimizer, with lr 3e-5, batch size 128, dropout 0.05, and Adam betas $(0.9, 0.95)$. We conduct early stopping, selecting the checkpoint with the lowest validation loss on a random held-out validation set, consisting of 10\% of the questions in the original PRM800k training split.

We finetune the PRM on 16 samples per question from the corresponding few-shot prompted base model. At each step, we use 16 monte-carlo rollouts, using the same base model and prompt, to estimate the step-level value. We filter out all samples which fail to output a valid, parsable final answer from the training data, as we found these to hurt PRM performance in initial experiments.

When generating the samples, the base model is prompted to output answers in newline separated a step-by-step format, as done in~\citet{lightman2023lets}. We then separate each of the answers into steps using a simple newline splitting procedure. We include details about our prompt in Appendix~\ref{app:prompting}.

\section{Comparing PRM Aggregation Strategies}
\label{app:comparing_prm_agg}

\begin{figure}
    \centering
    \includegraphics[width=0.6\textwidth]{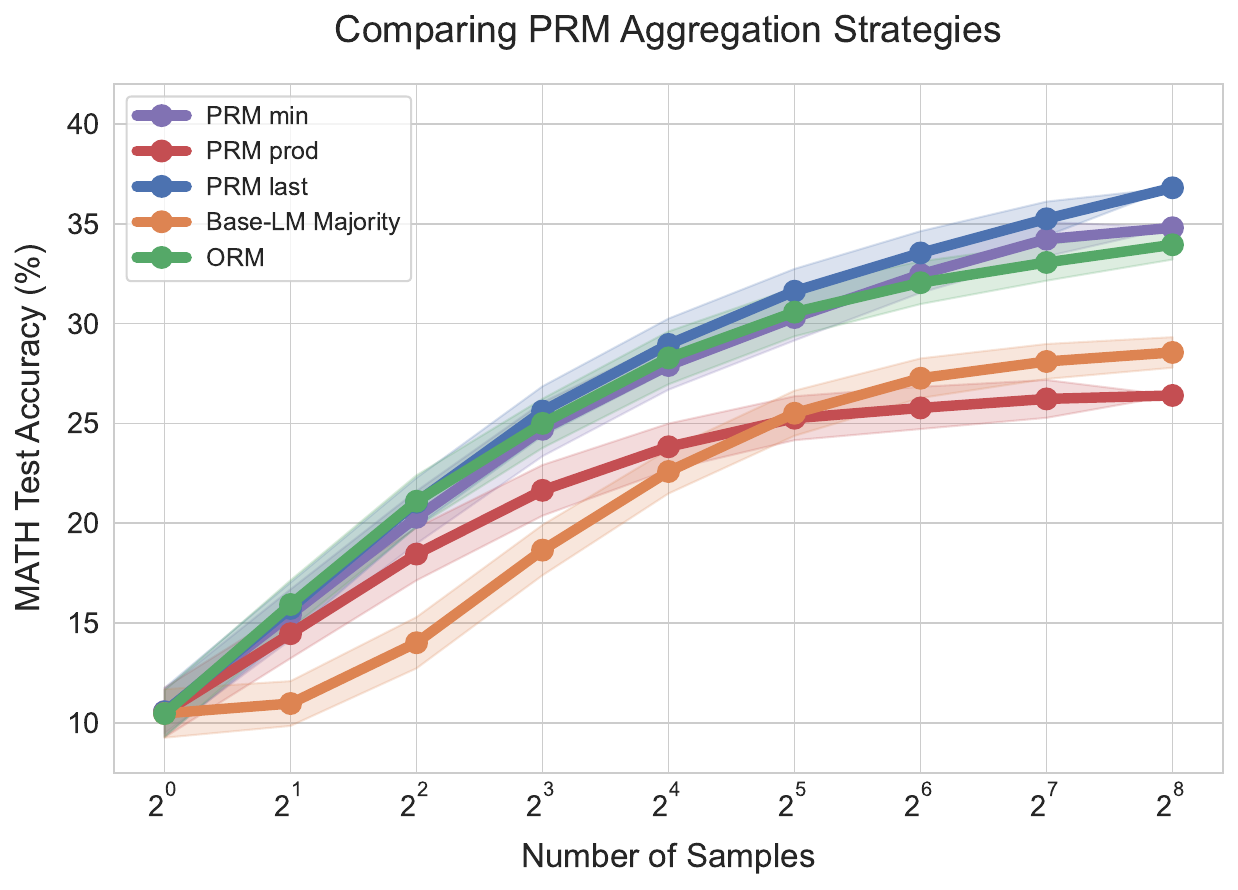}
    \caption{We compare different methods of aggregating per-step PRM scores to produce a final score for the full solution: ``min'' refers to taking the minimum score accross all steps, ``prod'' takes the product of all step correctness probabilities, and ``last'' just uses the last step score. We see that last performs the best across all aggregation strategies.}
    \label{fig:comparing_prm_aggregations}
\end{figure}

We compare different methods of aggregating per-step PRM scores to produce a final score for the full solution. Specifically we compare: 1) taking the minimum score accross all steps as done in~\citet{lightman2023lets} (e.g. ``min''); 2) taking the product of all step correctness probabilities (e.g. ``prod''); and 3) taking just the last step prediction (e.g. ``last''). We see in Figure~\ref{fig:comparing_prm_aggregations} that taking the last step outperforms the other two approaches. Prior works~\citep{lightman2023lets,wang2023mathshepherd} found min to be the best aggregator. We believe that the discrepancy is due to the fact that our verifier was trained with soft MC return labels, which surface very differently from binary correctness labels, and therefore other aggregation strategies may not have the same effect.

Interestingly, when using the last step aggregation, we are effectively using the PRM like an ORM. However, we see that the PRM outperforms the ORM, suggesting that in our case the per-step PRM training may be largely useful as a form of representation learning, rather than purely as a tool at inference time. Future work should further explore this line of reasoning.

\section{Comparing PRM and ORM}
\label{app:comparing_prm_orm}

\begin{figure}
    \centering
    \includegraphics[width=0.6\textwidth]{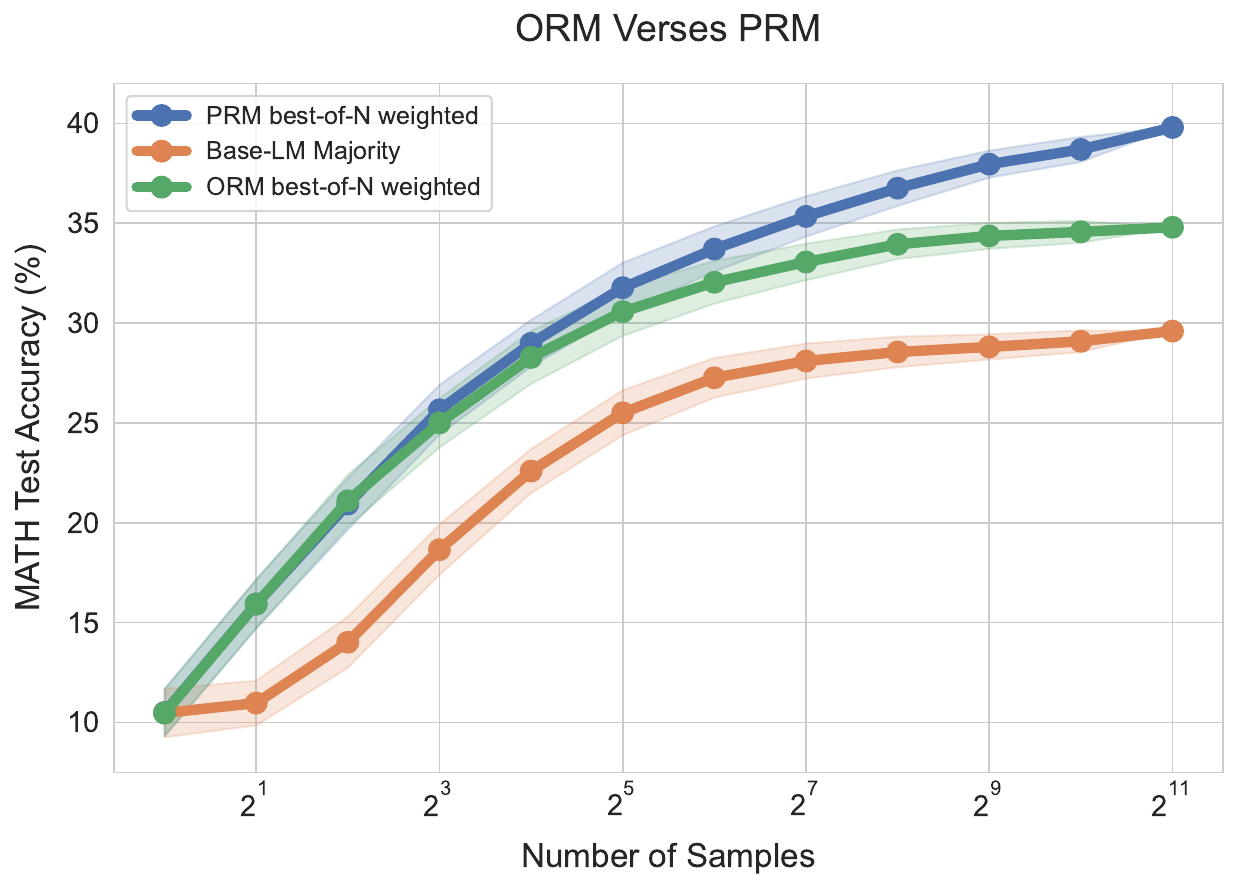}
    \caption{We compare PRM and ORM models finetuned from PaLM 2-S* in a best-of-N evaluation. We use the PaLM 2-S* base LM to sample outputs, using a few-shot prompt. We see that the PRM greatly outperforms the ORM at a larg number of samples.}
    \label{fig:prm_v_orm}
\end{figure}

We trained a PRM and ORM model using the PaLM 2-S* base LM. We see in Figure~\ref{fig:prm_v_orm}, that the PRM outperforms the ORM, and the gap between the gap between the PRM and ORM grows with the number of samples used. We use the last step prediction from the PRM to score the answers as described in Appendix~\ref{app:comparing_prm_agg}.

\section{Prompting Details}
\label{app:prompting}

In order to enable the base model to output answers in a step-by-step format to which a PRM can be applied, we use a 4-shot prompt consisting of randomly selected correct answer examples from the PRM800k data released by~\citet{lightman2023lets}. Specifically we use answers from the phase 1 training split. These answers correspond to GPT-4 generated correct answer examples, which include the correct step-by-step format. In initial experiments, we found that this prompting procedure produces similar results to the prompt used in~\citet{lewkowycz2022solving}. We use this prompt for generating training data for the PRM and the revision model. We also use this prompt when conducting search against the PRM on the test-set. To grade the final answer predicted by this prompt, we use the grading function released by~\citet{lightman2023lets}.

\section{Revision Model Finetuning Details}
\label{app:revision_finetune}

For fine-tuning the revision model, we follow the procedure outlined in Section~\ref{sec:revision_setup}. We first sample 64 outputs per question. We then filter out all answers which end in an invalid solution. For each correct answer, we then sample a number uniformly between 0 and 4 indicating how many incorrect answers to include in context for training. The correct answer is used as the last answer in the trajectory (which we train the model to produce) and the incorrect answers are included in context. If the sampled number is greater than 0, we then find the closest incorrect answer according to a character-level edit distance metric to include as the last incorrect answer in the trajectory. The goal here is to select an incorrect answer which is somewhat correlated with the correct answer, to improve learning. The remaining incorrect answers, we sample randomly from the set of available answers. In the case where there are fewer than 4 incorrect answers sampled, we truncate the uniform distribution's max to match the number of incorrect samples. We use this procedure to generate trajectories for all questions in the training data.

We then finetune the base language model on the correct answer solutions in these generated trajectories. We use the AdamW optimizer with lr 1e-5, batch size 128, dropout 0.0, and Adam betas $(0.9, 0.95)$.

We find that generally evaluating loss on an evaluation set consisting of trajectories generated as described above, does not provide a good signal for early stopping. Rather, we find that checkpoints much after the evaluation loss begins increasing are much more capable of revisions. This is likely because after finetuning the revision model, the evaluation set represents off-policy data, which will naturally be out-of-distribution compared to the trajectires that the model itself would generate on-policy. We therefore select our revision model checkpoint slightly after the point where we observe overfitting on the validation set.

\section{Revision Model Selection Criteria}
\label{app:revision_selection}

As described in Section~\ref{sec:revision_setup}, in order to effective use our revision model we need to deploy a criteria for selecting the best answer both within a revision trajectory and between multiple parallel trajectories. We use two approaches: 1) ORM verifier; and 2) majority voting.

For the ORM verifier, we train an ORM on the revision model's outputs according to the procedure in Appendix~\ref{app:revision_verifier}. At inference, time we then use this verifier to select the best answer. Since we have two axes across which to aggregate (within each revision trajectories and between multiple trajectories), we deploy a hierarchical strategy, first selecting the best answer within each revision trajectory and then aggregating these selected answers across trajectories. To select the best answer within each trajectory, we perform best-of-N weighted aggregation and then choose the highest scoring solution with the maximum best-of-N weighted answer. Then, to select the final answer across all revision chains, we perform another round of best-of-N weighted selection using the best answer from each revision chain. The answer after this second round of best-of-N weighted represents our final answer prediction.

For majority voting we found hierarchical aggregation to create problems when the length of the trajectory or the number of trajectories was too small. The problem being that without enough samples, majority voting is unable to effectively select the best option. Therefore, for majority voting, we simply take all answers, across all trajectories, at once and take their majority as the final-answer. We found this to produce much smoother scaling behavior than the hierarchical approach.

\section{Revision Model Verifier Training}
\label{app:revision_verifier}

\begin{figure}
    \centering
    \subfigure{
        \includegraphics[width=0.48\textwidth]{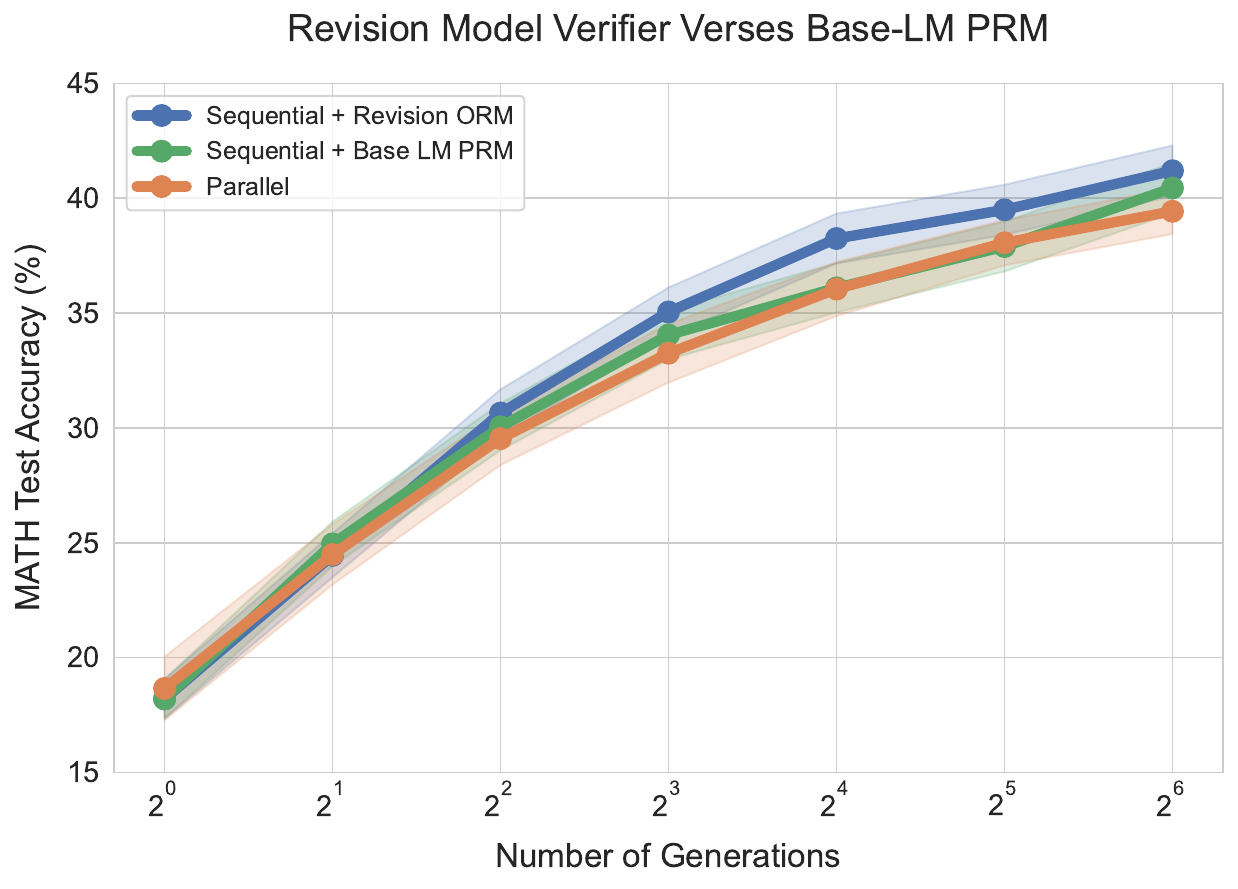}
        \label{fig:revision_verifier_v_prm}
    }\hfill
    \subfigure{
        \includegraphics[width=0.48\textwidth]{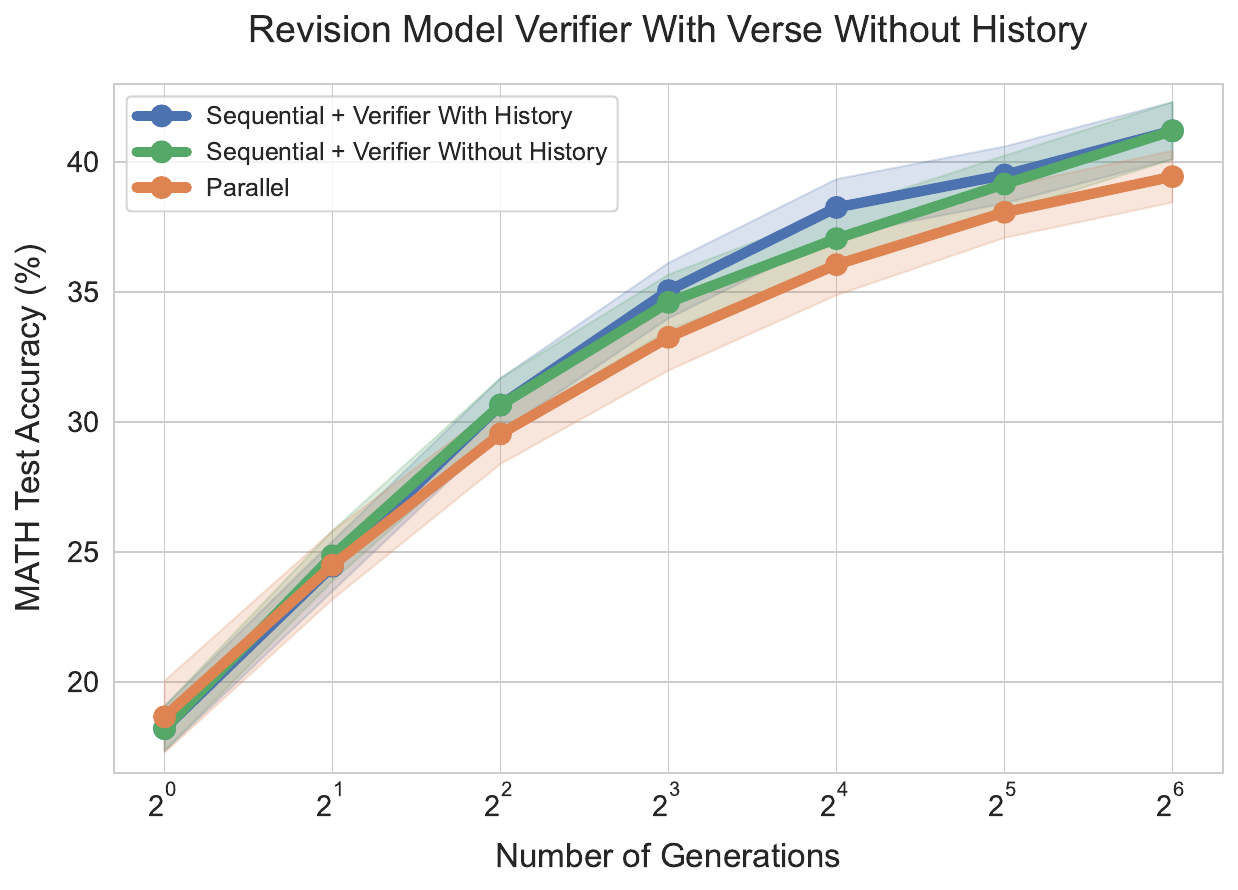}
        \label{fig:revision_verifier_w_v_wo_history}
    }
    \caption{Left: we compare the ORM we trained on the revision model's outputs against the PRM we trained on the PaLM 2-S* base model's outputs. We see that when applied to outputs from the revision model, the ORM adapted to the revision model outperforms the PRM, likely due to distribution shift with the revision model. Right: we ablate the effect of including previous revisions in the revision model verifier's context. We see that including revisions in-context helps the verifier slightly, but both settings still outperform the parallel baseline.}
    \label{fig:revision_verifier_ablations}
\end{figure}

We found that the PRM we finetuned on the PaLM 2-S* base model outputs was not as effective when applied to the PaLM 2-S* revision model's outputs (see Figure~\ref{fig:revision_verifier_v_prm}), likely due to distribution shift with the revision model. We therefore, trained a separate ORM verifier to use with our PaLM 2-S* revision model. We could have trained a PRM as well, but opted for an ORM due to the high cost of generating per-step PRM labels.

We modified the standard ORM slightly for the revision setting, by finetuning the ORM with previous revision in context, such that the verifier has access to the same context as the revision model, allowing the verifier see the revision model's previous answer attempts when scoring the current answer. All other experiment details are identical to those used for training the PRM.

Empirically, we find that including the revision history in context improves performance slightly (see Figure~\ref{fig:revision_verifier_w_v_wo_history}). Additionally, even without the revisions in context, we see that sequential revisions still slightly outperforms parallel, demonstrating improvements from sequential sampling are not just due to the verifier's context.

\section{$\text{ReST}^{\text{EM}}$ Revision Model Experiments}
\label{app:rest_revision_model}

We experimented with further optimizing our PaLM 2-S* revision model by training the model with a simplified RL algorithm: $\text{ReST}^{\text{EM}}$~\citep{singh2024human}. Specifically, we generated 64 revision trajectories of maximum length 5 for each question on the MATH training set. We stopped the revision model at the first correct answer in each trajectory. Using this generated data, we then finetuned the base LM on the correct answer data. To help the model learn the task, we explicitly balanced the distribution of trajectory lengths.

\begin{figure}
    \centering
    \includegraphics[width=0.6\textwidth]{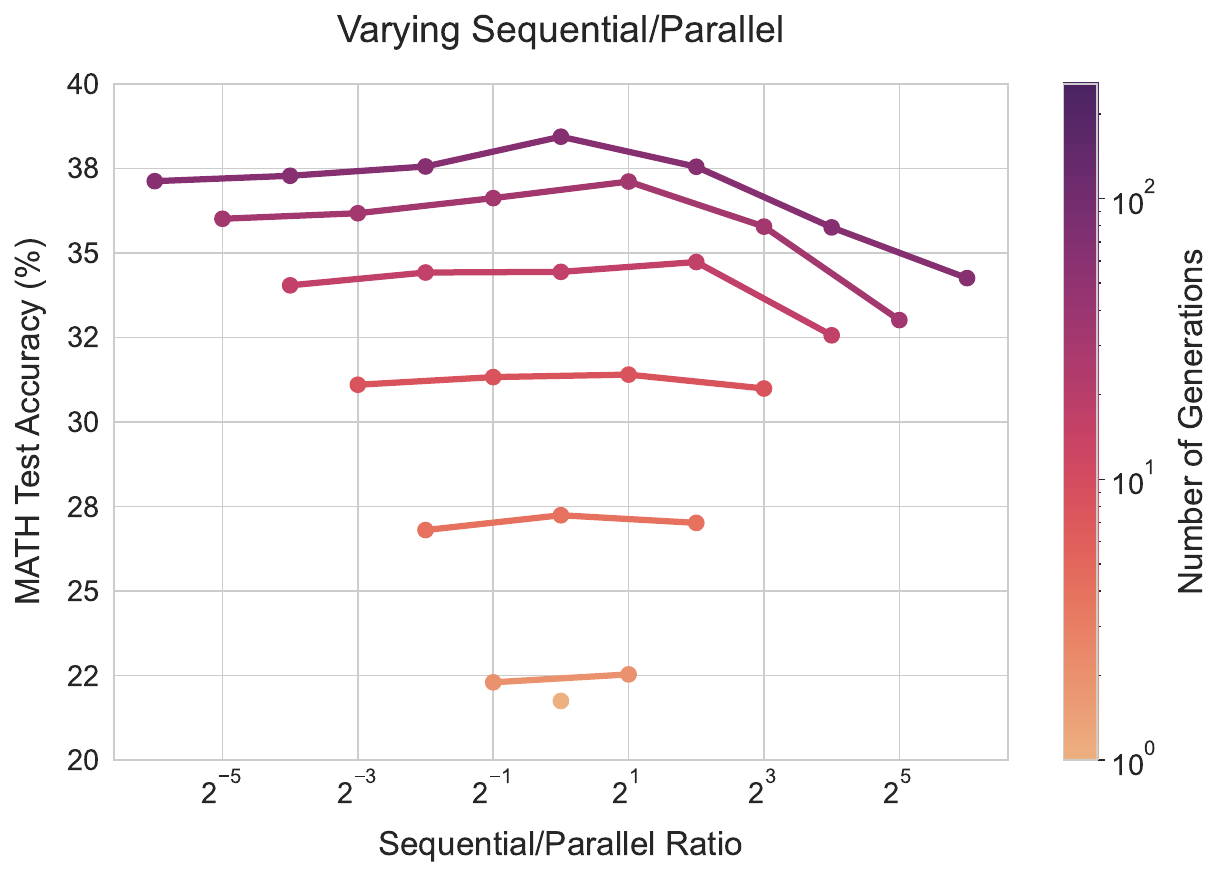}
    \caption{Performance of our $\text{ReST}^{\text{EM}}$ optimized revision model as the sequential to parallel ratio is varied. We use majority voting to select the answer. We see that this optimized revision model demonstrates substantial performance degradations with additional sequential revisions.}
    \label{fig:rest_em_revisions}
\end{figure}

In Figure~\ref{fig:rest_em_revisions}, we plot the performance of this new revision model as we vary the sequential to parallel ratio. We see that additional sequential revisions substantially hurts performance with this new model. We hypothesize that this degradation is due to the fact that the online data obtained from running $\text{ReST}^{\text{EM}}$ exacerbates spurious correlations in revision data, causing the optimized model to fail to learn the revision task. We believe that using a more offline data collection strategy, as done in~\citet{qu2024recursive}, may be more effective, and leave further exploration to future work.

\section{Revision Model Example Outputs}
\label{app:revision_example_outputs}

\begin{figure}
    \centering
    \includegraphics[width=0.75\textwidth]{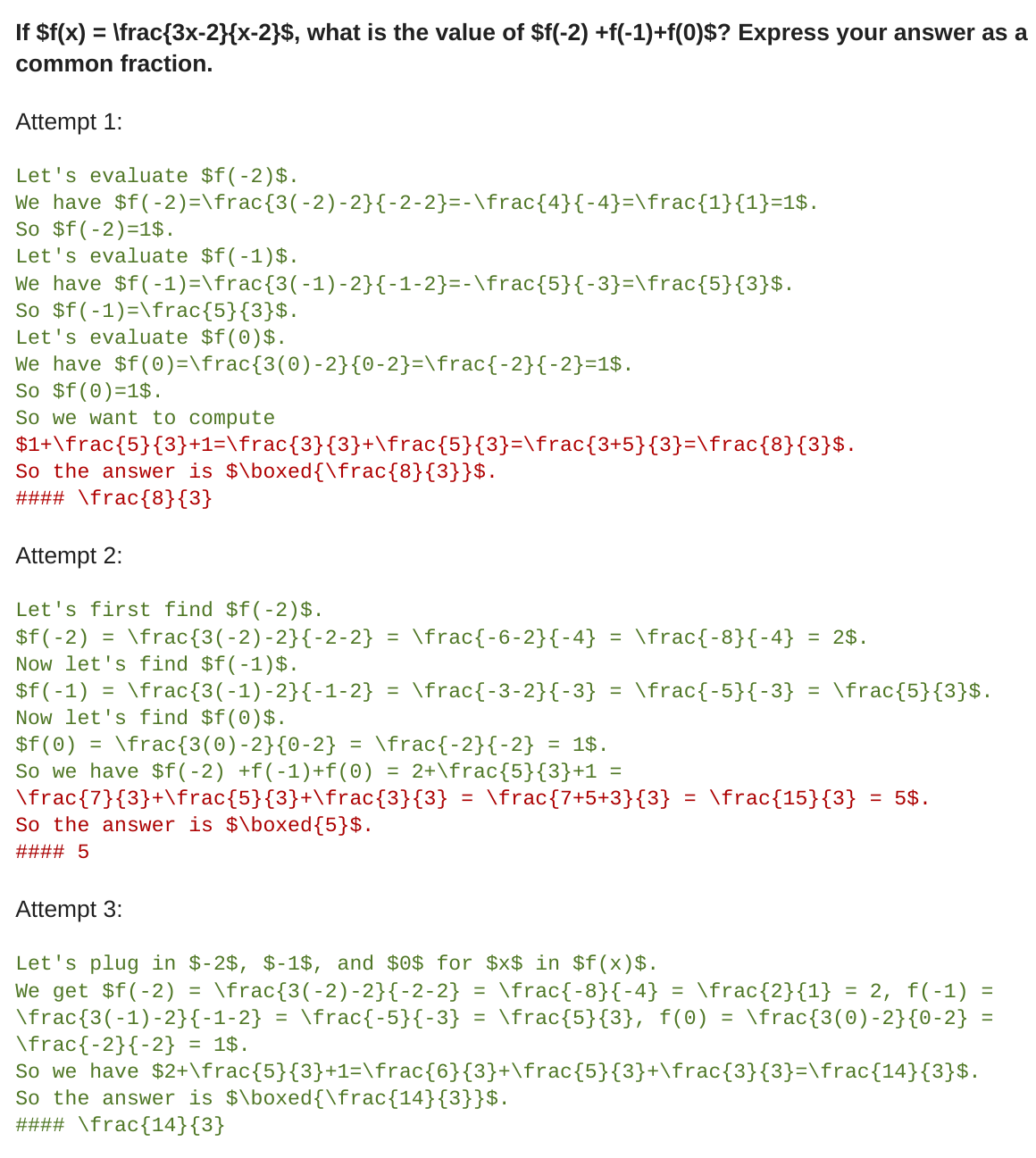}
    \caption{Revision model example 1. The model calculates the sum at the end incorrectly on the first two attempts, but on the third attempt it succeeds and gets the answer correct.}
    \label{fig:revisions_ex1}
\end{figure}

\begin{figure}
    \centering
    \includegraphics[width=0.75\textwidth]{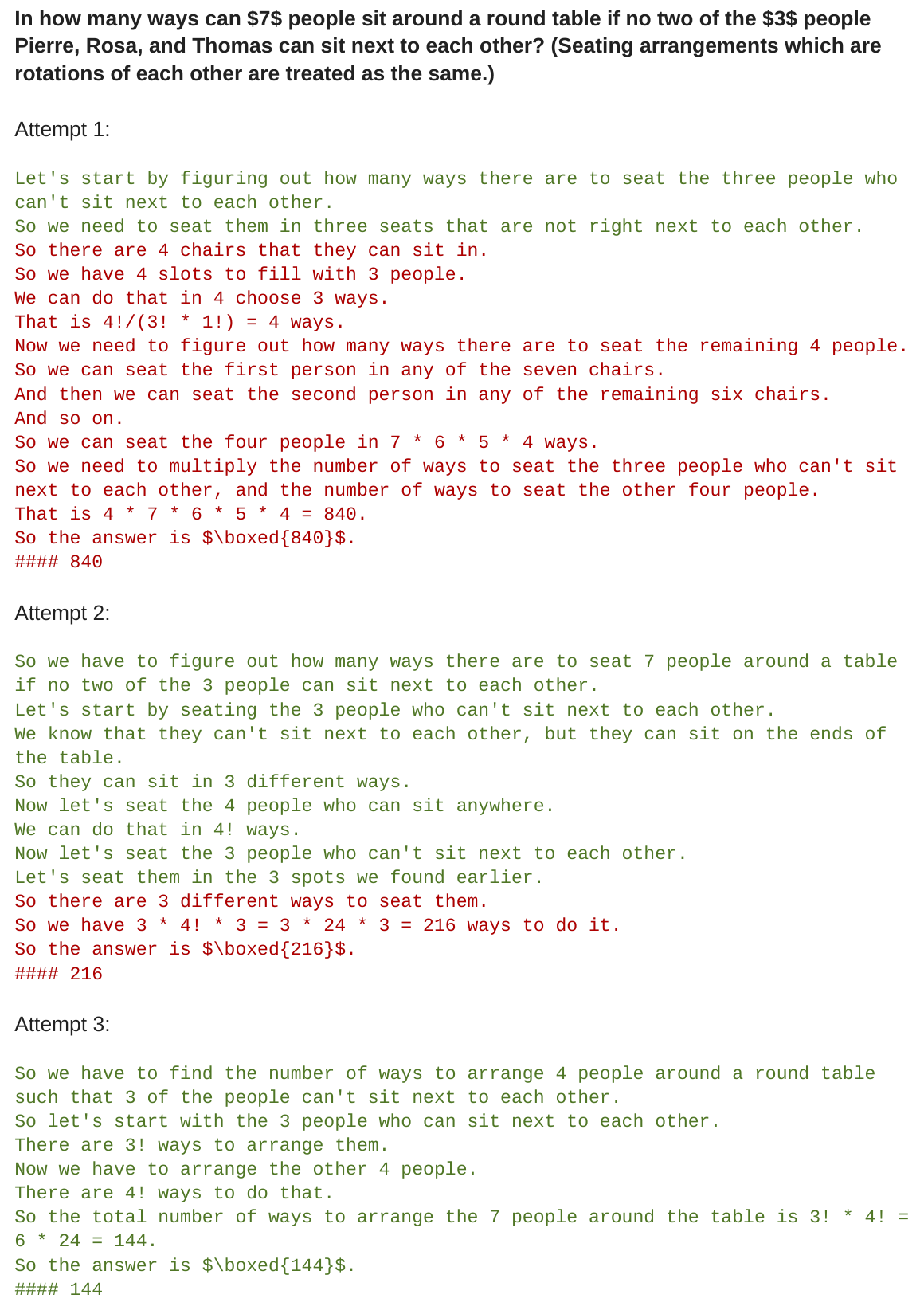}
    \caption{Revision model example 2. On the first attempt the model takes the incorrect approach, on the second attempt it gets closer but then makes a mistake towards the end. On the final attempt it gets to the correct answer.}
    \label{fig:revisions_ex2}
\end{figure}

\begin{figure}
    \centering
    \includegraphics[width=0.75\textwidth]{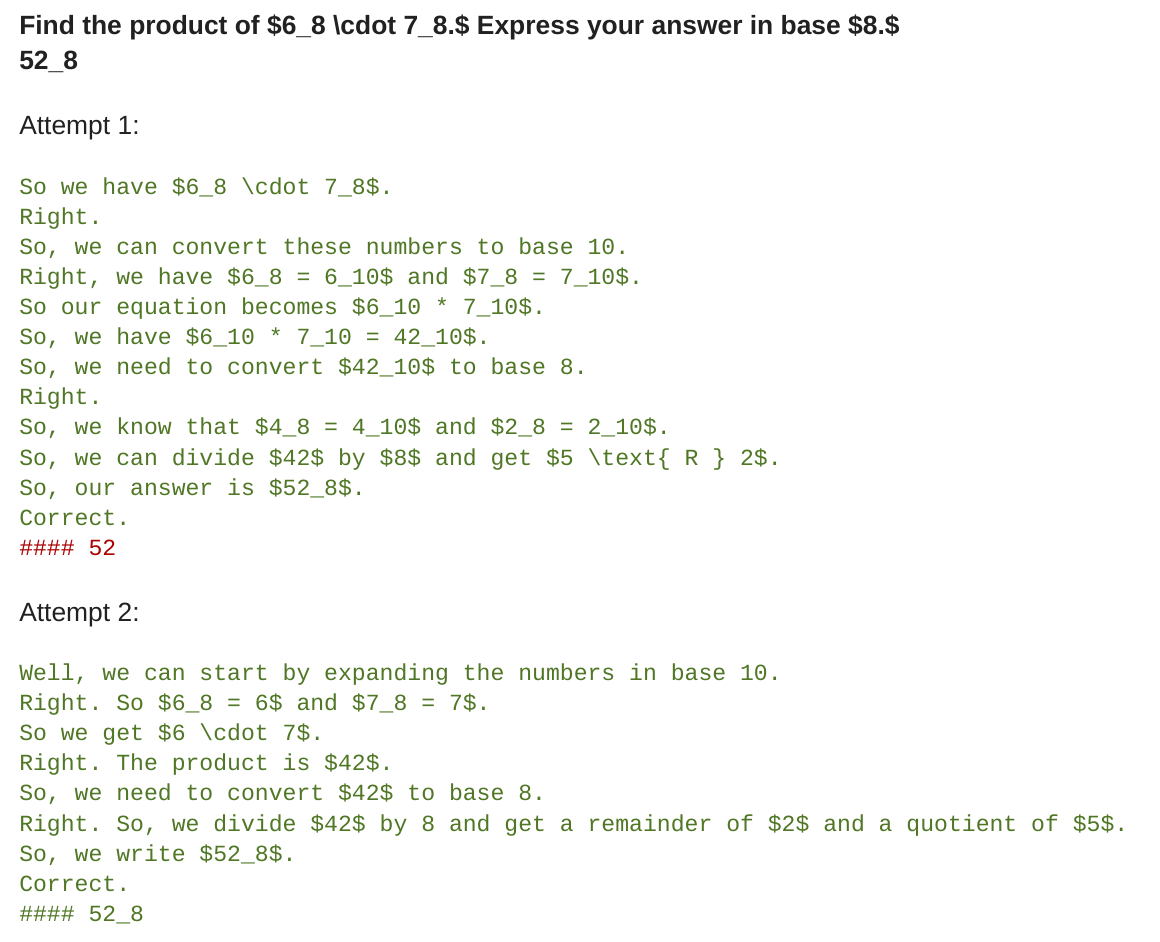}
    \caption{Revision model example 3. On the first attempt the model makes a mistake with the formatting of the final answer; it corrects this on the second attempt.}
    \label{fig:revisions_ex3}
\end{figure}

\begin{figure}
    \centering
    \includegraphics[width=0.75\textwidth]{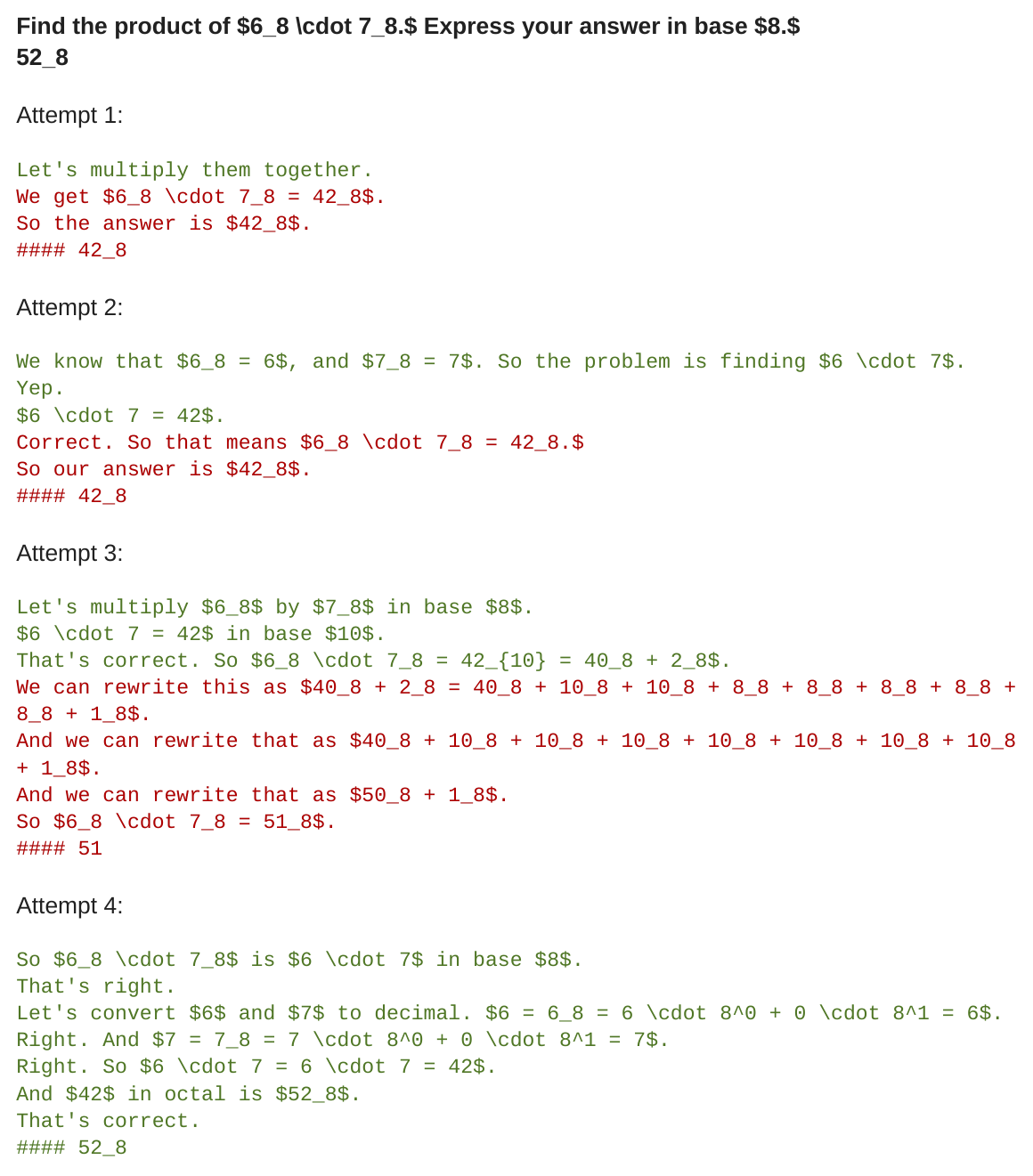}
    \caption{Revision model example 4. On the first few attempts the model fails the base 10 to base 8 conversion. On the final attempt it makes the correct calculation.}
    \label{fig:revisions_ex4}
\end{figure}

\begin{figure}
    \centering
    \includegraphics[width=0.75\textwidth]{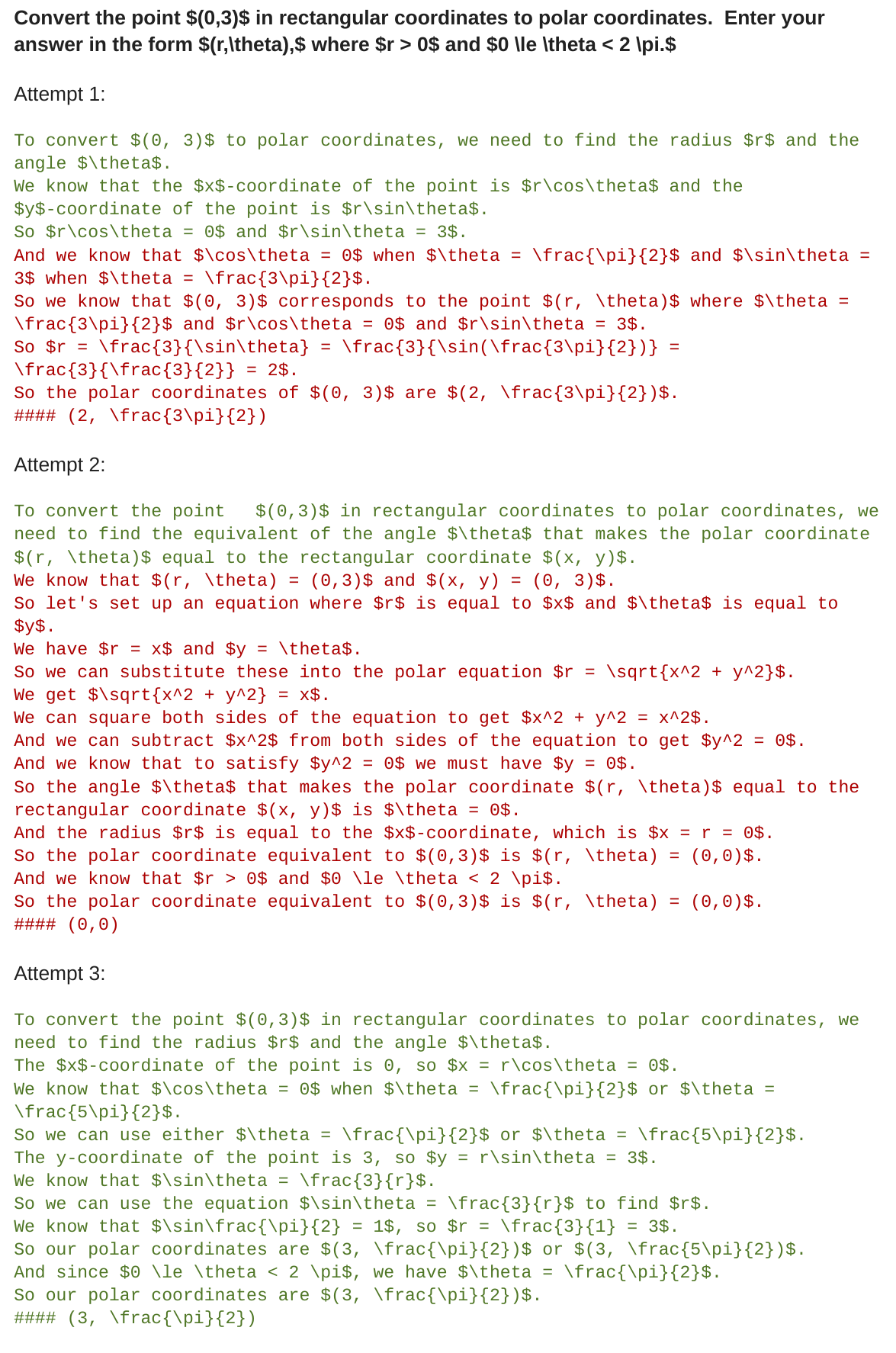}
    \caption{Revision model example 5. On the first two attempts the model makes an error when converting euclidean to polar coordinates. On the final attempt it does not make these mistakes.}
    \label{fig:revisions_ex5}
\end{figure}

\begin{figure}
    \centering
    \includegraphics[width=0.75\textwidth]{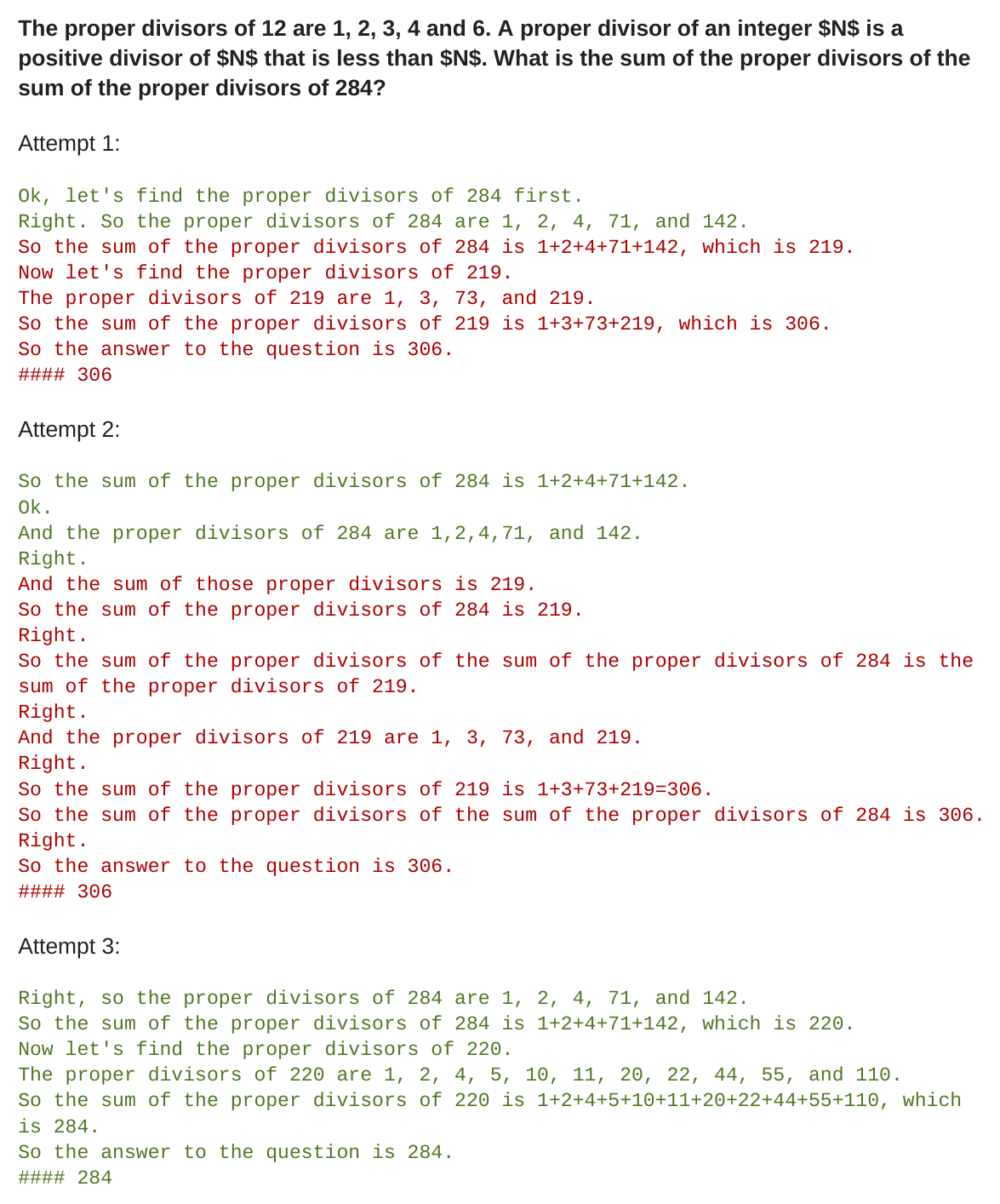}
    \caption{Revision model example 6. On the first two attempts the model makes a mistake when summing the proper divisors of 284. On the third attempt, it evaluates this sum correctly.}
    \label{fig:revisions_ex6}
\end{figure}

\begin{figure}
    \centering
    \includegraphics[width=0.75\textwidth]{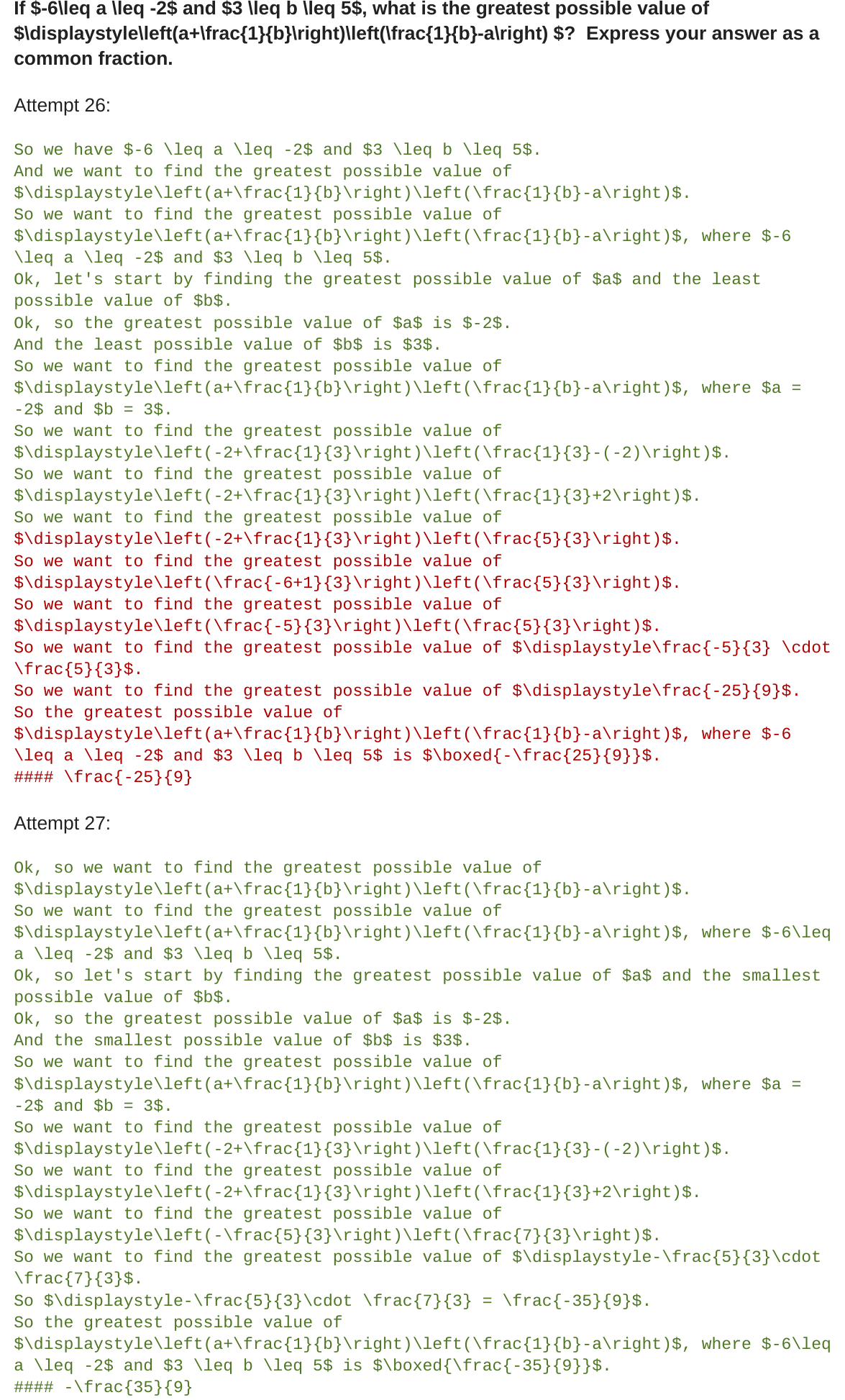}
    \caption{Revision model example 7. On the first attempt the model evaluates $\frac{1}{3}+2$ incorrectly. On the second attempt it corrects this error.}
    \label{fig:revisions_ex7}
\end{figure}

In Figures~\ref{fig:revisions_ex1},~\ref{fig:revisions_ex2},~\ref{fig:revisions_ex3},~\ref{fig:revisions_ex4},~\ref{fig:revisions_ex5},~\ref{fig:revisions_ex6}, and~\ref{fig:revisions_ex7}, we include select examples of our revision model's outputs.

\section{PRM Beam Search Example Outputs}
\label{app:prm_example_outputs}

\begin{figure}
    \centering
    \includegraphics[width=1.0\textwidth]{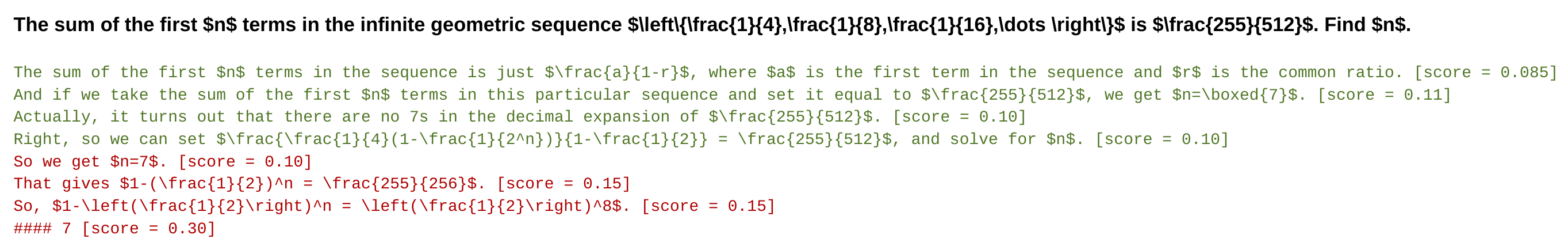}
    \caption{PRM beam search example 1.}
    \label{fig:prm_ex1}
\end{figure}

\begin{figure}
    \centering
    \includegraphics[width=1.0\textwidth]{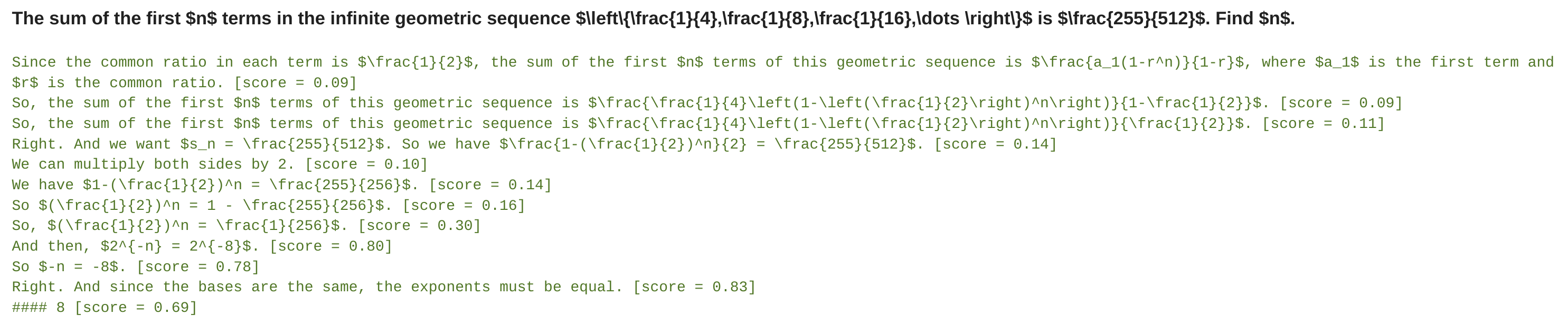}
    \caption{PRM beam search example 2.}
    \label{fig:prm_ex2}
\end{figure}

\begin{figure}
    \includegraphics[width=0.33\textwidth]{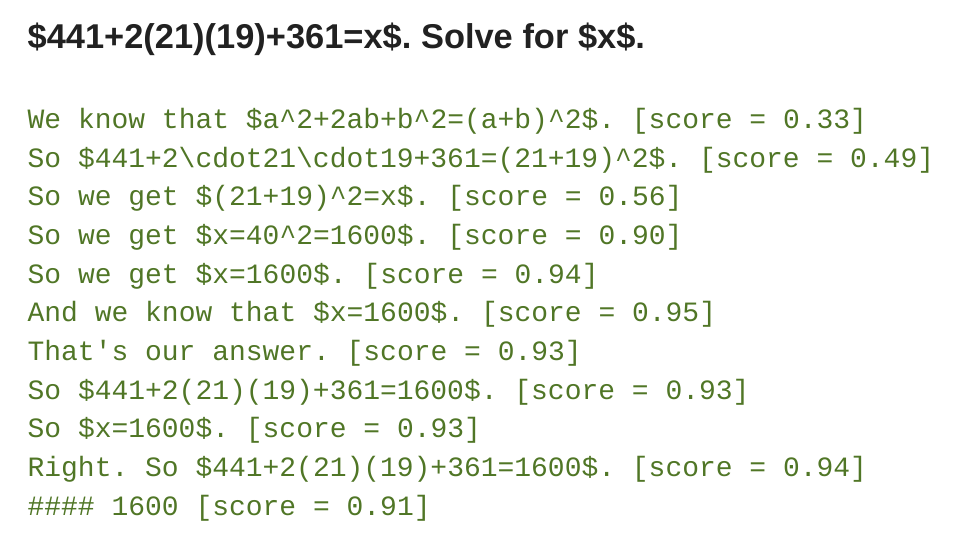}
    \caption{PRM beam search example 3.}
    \label{fig:prm_ex3}
\end{figure}

\begin{figure}
    \centering
    \includegraphics[width=1.0\textwidth]{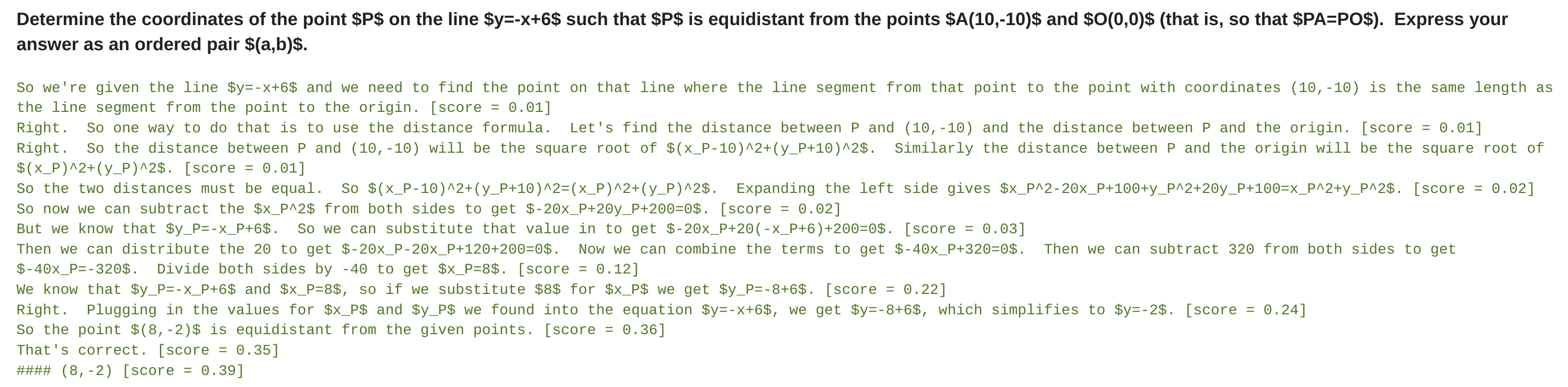}
    \caption{PRM beam search example 4.}
    \label{fig:prm_ex4}
\end{figure}

\begin{figure}
    \centering
    \includegraphics[width=1.0\textwidth]{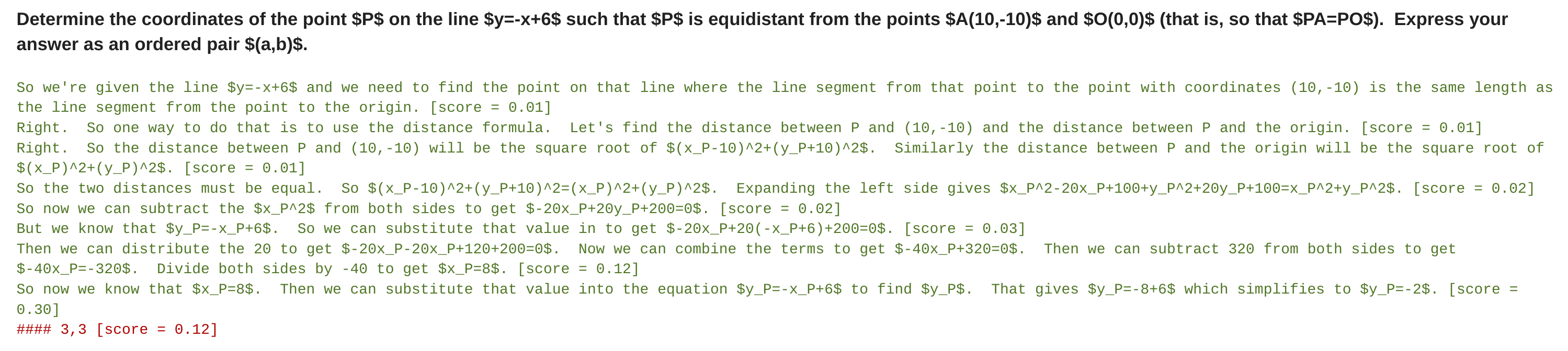}
    \caption{PRM beam search example 5.}
    \label{fig:prm_ex5}
\end{figure}

\begin{figure}
    \includegraphics[width=0.6\textwidth]{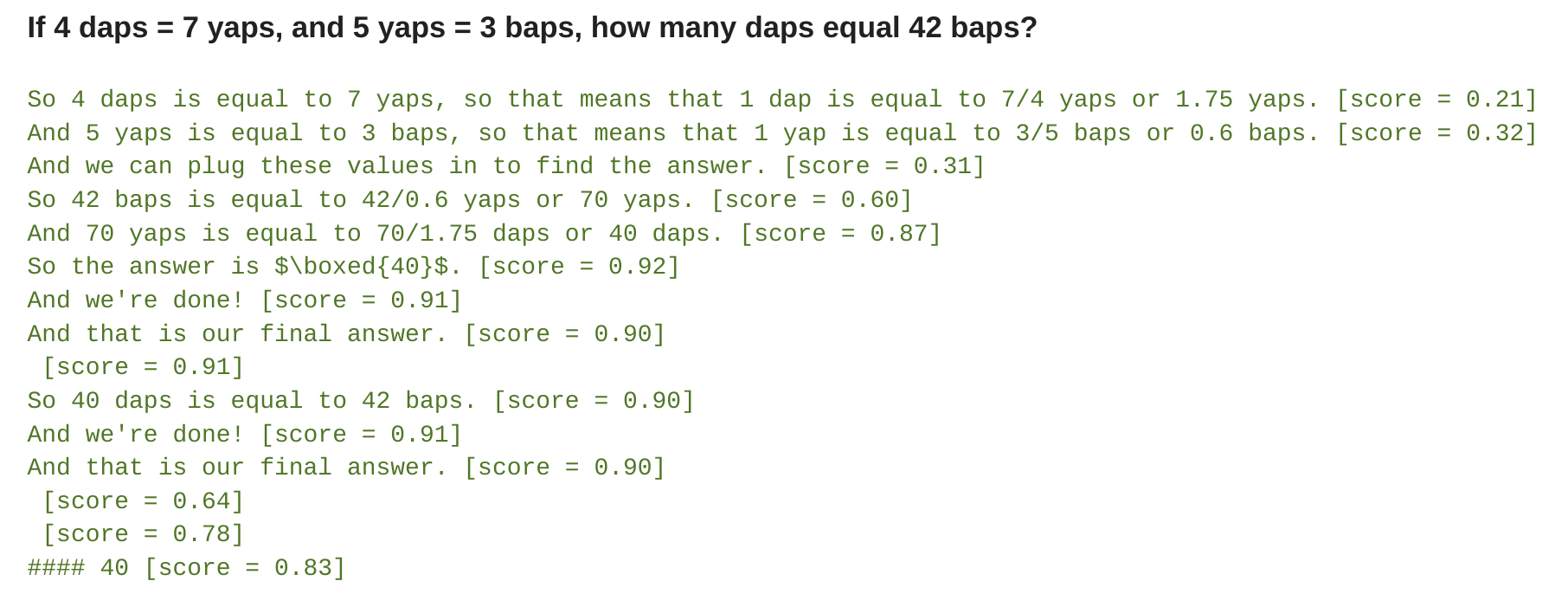}
    \caption{PRM beam search example 6.}
    \label{fig:prm_ex6}
\end{figure}

In Figures~\ref{fig:prm_ex1},~\ref{fig:prm_ex2},~\ref{fig:prm_ex3},~\ref{fig:prm_ex4},~\ref{fig:prm_ex5}, and~\ref{fig:prm_ex6}, we include select examples of PRM beam search. We include the PRM score, between 0 and 1, for each step in the examples.

\end{document}

%% file: template_content.tex
\newcommand{\expect}[2]{\mathds{E}_{{#1}} \left[ {#2} \right]}
\newcommand{\myvec}[1]{\boldsymbol{#1}}
\newcommand{\myvecsym}[1]{\boldsymbol{#1}}
\newcommand{\vx}{\myvec{x}}
\newcommand{\vy}{\myvec{y}}
\newcommand{\vz}{\myvec{z}}
\newcommand{\vtheta}{\myvecsym{\theta}}

\vspace{-0.25cm}
\section{Introduction}
\vspace{-0.2cm}

Humans tend to think for longer on difficult problems to reliably improve their decisions~\citep{kahneman2013thinking,evans1984heuristic,kahneman2003maps}.
Can we instill a similar capability into today's large language models (LLMs)?  More specifically, given a challenging input query, \emph{can we enable language models to most effectively make use of additional computation at test time so as to improve the accuracy of their response?} In theory, by applying additional computation at test time, an LLM should be able to do better than what it was trained to do. In addition, such a capability at test-time also has the potential to unlock new avenues in agentic and reasoning tasks~\citep{shinn2023reflexion,wei2023chainofthought,qu2024recursive}.
For instance, if pre-trained model size can be traded off for additional computation during inference, this would enable LLM deployment in use-cases where smaller on-device models could be used in place of datacenter scale LLMs.
Automating the generation of improved model outputs by using additional inference-time computation also provides a path towards a general self-improvement algorithm that can function with reduced human supervision.

\begin{figure}[t]
    \centering
    \subfigure{
        \includegraphics[width=0.85\textwidth]{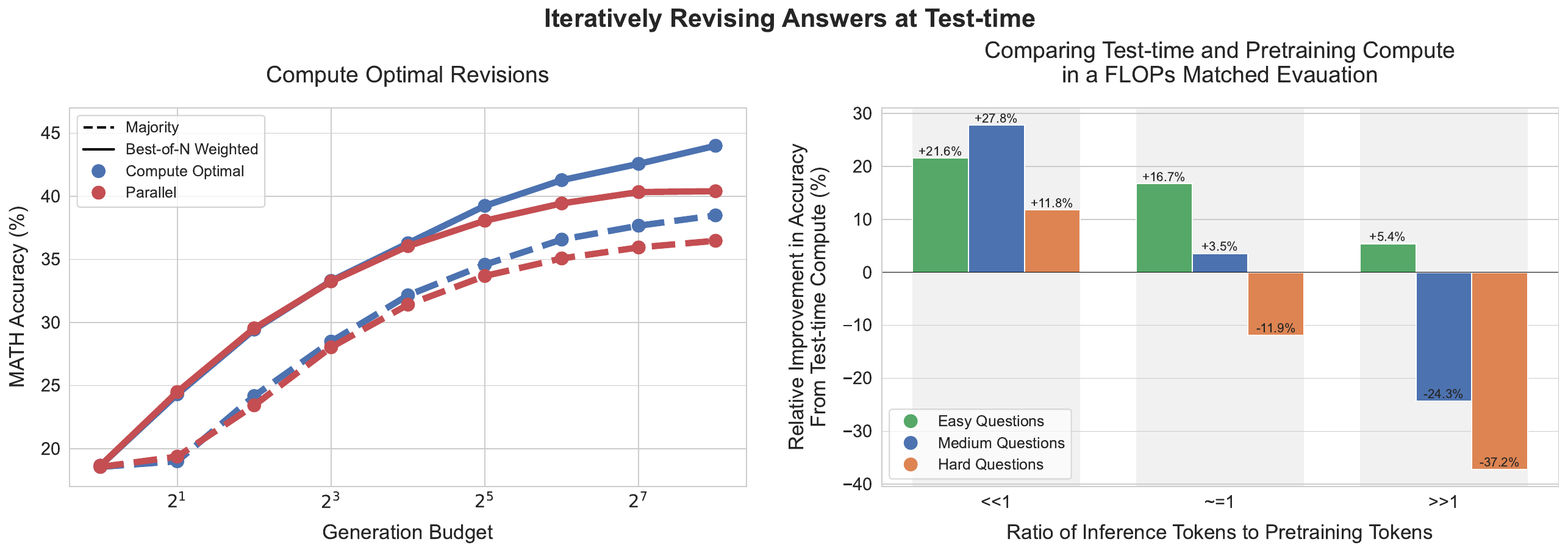}
    } \\
    \subfigure{
        \includegraphics[width=0.85\textwidth]{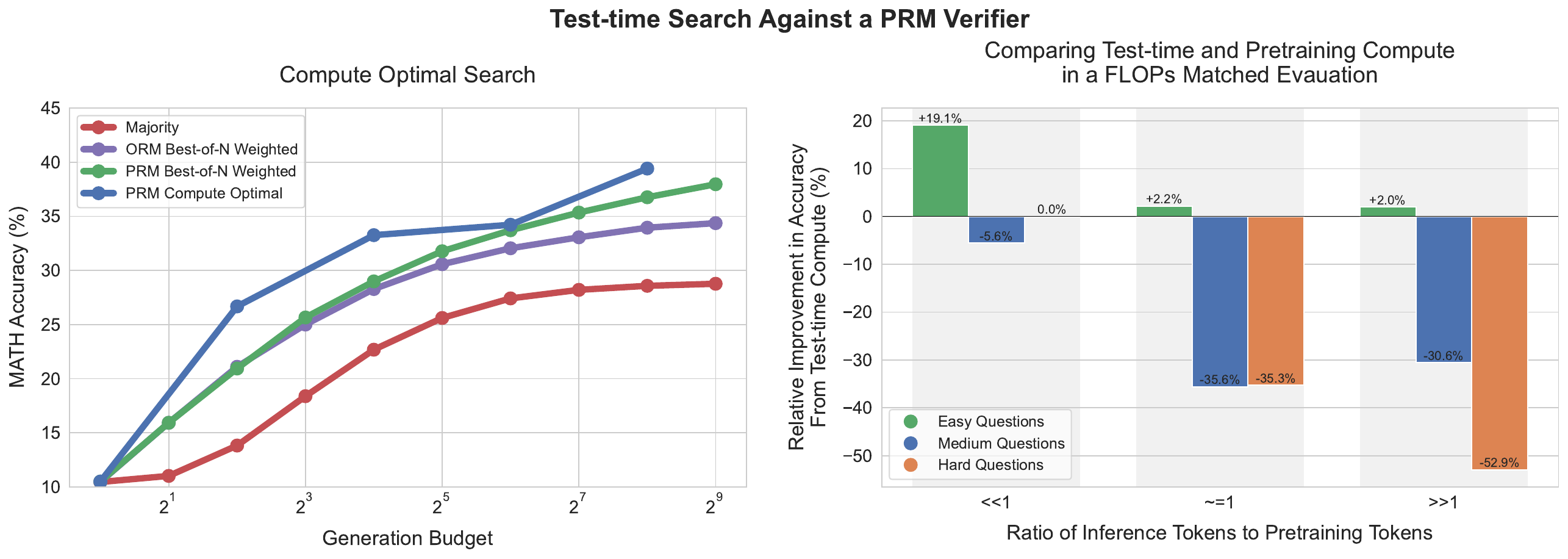}
    }
    \vspace{-0.2cm}
    \caption{\scriptsize{\textbf{\emph{Summary of our main results.}} \textbf{Left: Compute-optimal scaling for iterative self-refinement (i.e., revisions) and search.} On the left, we compare the compute-optimal scaling policy for our PaLM 2-S* revision model against baselines in the revision setting (top) and the PRM search setting (bottom). We see that in the revisions case, the gap between standard best-of-N (e.g. ``parallel'') and compute-optimal scaling gradually widens, enabling compute-optimal scaling to outperform best-of-N with $4\times$ less test-time compute. Similarly, in the PRM search setting, we observe significant early improvements over best-of-N from compute-optimal scaling, nearly outperforming best-of-N with $4\times$ less compute at points. See Sections~\ref{sec:search} and~\ref{sec:revisions} for details. \textbf{Right: Comparing test-time compute and model parameter scaling.} We compare the performance of compute-optimal test-time scaling with PaLM 2-S* against the performance of a $\sim14\times$ larger pretrained model without additional test-time compute (e.g. greedy sampling). We consider the setting where we expect $X$ tokens of pretraining for both models and $Y$ tokens of inference. By training a larger model, we effectively multiply the FLOPs requirement for both of these terms. If we were to apply additional test-time compute with the smaller model, so as to match this larger model's FLOPs requirement, how would it compare in terms of accuracy? We see that for the revisions (top) when $Y << X$, test-time compute is often preferable to additional pretraining. However, as the inference to pretraining token ratio increases, test-time compute remains preferable on easy questions. Whereas on harder questions, pretraining is preferable in these settings. We also see a similar trend with PRM search (bottom). See Section~\ref{sec:exchanging} for more details.}}
    \vspace{-0.2cm}
    \label{fig:comparing_base_lms}
\end{figure}

Prior work studying inference-time computation provides mixed results. On the one hand, some works show that current LLMs can use test-time computation to improve their outputs~\citep{bai2022constitutional,madaan2023selfrefine,du2023improving,saunders2022selfcritiquing,yao2023tree}, on the other hand, other work shows that the effectiveness of these methods on more complex tasks such as math reasoning remains highly limited~\citep{huang2023large,stechly2023gpt4,valmeekam2023large}, even though reasoning problems often require drawing inferences about existing knowledge as opposed to new knowledge. 
These sorts of conflicting findings motivate the need for a systematic analysis of different approaches for scaling test-time compute.

We are interested in understanding the benefits of scaling up test-time compute. Arguably the simplest and most well-studied approach for scaling test-time computation is \emph{best-of-N} sampling: sampling N outputs in ``parallel'' from a  base LLM and selecting the one that scores the highest per a learned verifier or a reward model~\citep{cobbe2021training,lightman2023lets}. However, this approach is not the only way to use test-time compute to improve LLMs. By modifying either the \emph{proposal distribution} from which responses are obtained (for instance, by asking the base model to revise its original responses ``sequentially''~\citep{qu2024recursive}) or by altering how the \emph{verifier} is used (e.g. by training a process-based dense verifier~\citep{lightman2023lets,wang2023mathshepherd} and searching against this verifier), the ability scale test-time compute could be greatly improved, as we show in the paper. 

To understand the benefits of scaling up test-time computation, we carry out experiments on the challenging MATH~\citep{hendrycks2021measuring} benchmark using PaLM-2~\citep{anil2023palm} models specifically fine-tuned\footnote{Capability-specific finetuning is necessary to induce revision and verification capabilities into the base model on MATH since these capabilities are absent even in strong proprietary LLMs~\citep{huang2023large,sharma2024criticalevaluationaifeedback}. However, we expect that future LLMs will be more effective at verification and revision due to both increased scale and the inclusion of additional data targeted specifically towards these capabilities~\citep{predictemergence,blakeney2024doesdatasparkjoy,llmcriticscatchbugs}. Therefore in order to make progress towards understanding scaling of test-time computation, we must use models finetuned for these capabilities. That said, we expect future models to be pretrained for such capabilities directly, therefore avoiding the need for capability-specific finetuning.} to either revise incorrect answers~\citep{qu2024recursive} (e.g. improving the proposal distribution; Section~\ref{sec:revisions}) or verify the correctness of individual steps in an answer using a process-based reward model (PRM)~\citep{lightman2023lets,wang2023mathshepherd} (Section~\ref{sec:search}).
With both approaches, we find that the efficacy of a particular test-time compute strategy depends critically on both the nature of the specific problem at hand and the base LLM used. For example, on easier problems, for which the base LLM can already readily produce reasonable responses, allowing the model to iteratively refine its initial answer by predicting a sequence of N revisions (i.e., modifying the proposal distribution), may be a more effective use of test-time compute than sampling N independent responses in parallel. On the other hand, with more difficult problems that may require searching over many different high-level approaches to solving the problem, re-sampling new responses independently in parallel or deploying tree-search against a process-based reward model is likely a more effective way to use test-time computation.
This finding illustrates \textbf{the need to deploy an adaptive {``compute-optimal``} strategy for scaling test-time compute}, wherein the specific approach for utilizing test-time compute is selected depending on the prompt, so as to make the best use of additional computation.
We also show that a notion of \emph{question difficulty} (Section~\ref{sec:setup}) from the perspective of the base LLM can be used to predict the efficacy of test-time computation, enabling us to practically instantiate this `compute-optimal' strategy given a prompt. By appropriately allocating test-time compute in this way, we are able to greatly improve test-time compute scaling, surpassing the performance of a best-of-N baseline while only using about \textbf{4x} less computation with both revisions and search (Sections~\ref{sec:search} and~\ref{sec:revisions}).

Using our improved test-time compute scaling strategy, we then aim to understand to what extent test-time computation can effectively substitute for additional pretraining. We conduct a FLOPs-matched comparison between a smaller model with additional test-time compute and pretraining a \textbf{14x larger model}. We find that on easy and intermediate questions, and even hard questions (depending on the specific conditions on the pretraining and inference workload), additional test-time compute is often preferable to scaling pretraining. This finding suggests that rather than focusing purely on scaling pretraining, \textbf{in some settings it is be more effective to pretrain smaller models with less compute, and then apply test-time compute to improve model outputs}. That said, with the most challenging questions, we observe very little benefits from scaling up test-time compute. Instead, we find that on these questions, it is more effective to make progress by applying additional pretraining compute, demonstrating that current approaches to scaling test-time compute may not be 1-to-1 exchangeable with scaling pretraining. Overall, this suggests that even with a fairly na\"ive methodology, scaling up test-time computation can already serve to be more preferable to scaling up pretraining, with only more improvements to be attained as test-time strategies mature. Longer term, this hints at a future where fewer FLOPs are spent during pretraining and more FLOPs are spent at inference.

\vspace{-0.2cm}
\section{A Unified Perspective on Test-Time Computation: Proposer and Verifier}
\label{sec:unifying_perspective}
\vspace{-0.2cm}

We first unify approaches for using test-time computation and then analyze some representative methods. First, we view the use of additional test-time compute through the lens of modifying the model's predicted distribution \emph{adaptively} at test-time, conditioned on a given prompt. Ideally, test-time compute should modify the distribution so as to generate better outputs than na\"ively sampling from the LLM itself would. In general, there are two knobs to induce modifications to an LLM's distribution: \textbf{(1)} \textbf{at the input level}: by augmenting the given prompt with an additional set of tokens that the LLM conditions on to obtain the modified distribution, or \textbf{(2)} \textbf{at the output level}: by sampling multiple candidates from the standard LM and performing surgery on these candidates. In other words, we could either modify the \textbf{proposal distribution} induced by the LLM itself such that it is an improvement over na\"ively conditioning on the prompt or we could use some post-hoc \textbf{verifiers or scorers} to perform output modifications. This process is reminiscent of Markov chain Monte Carlo (MCMC)~\citep{andrieu2003introduction} sampling from a complex target distribution but by combining a simple proposal distribution and a score function.
Modifying the proposal distribution directly by altering input tokens and using a verifier form two independent axes of our study.

\textbf{Modifying the proposal distribution.} One way to improve the proposal distribution is to directly optimize the model for a given reasoning task via RL-inspired finetuning methods such as STaR or $\text{ReST}^{\text{EM}}$~\citep{zelikman2022star,singh2024human}. Note that these techniques do not utilize any additional input tokens but specifically finetune the model to induce an improved proposal distribution.
Instead, techniques such as self-critique~\citep{bai2022constitutional,madaan2023selfrefine,du2023improving,saunders2022selfcritiquing} enable the model itself to improve its own proposal distribution at test time by instructing it to critique and revise its own outputs in an iterative fashion. Since prompting off-the-shelf models is not effective at enabling effective revisions at test time, we specifically finetune models to iteratively revise their answers in complex reasoning-based settings. To do so, we utilize the approach of finetuning on on-policy data with Best-of-N guided improvements to the model response~\citep{qu2024recursive}. 

\textbf{Optimizing the verifier.} In our abstraction of the proposal distribution and verifier, the verifier is used to aggregate or select the best answer from the proposal distribution. The most canonical way to use such a verifier is by applying best-of-N sampling, wherein we sample N complete solutions and then select the best one according to a verifier~\citep{cobbe2021training}. However, this approach can be further improved by training a process-based verifier~\citep{lightman2023lets}, or a process reward model (PRM), which produces a prediction of the correctness of each intermediate step in an solution, rather than just the final answer. We can then utilize these per-step predictions to perform tree search over the space of solutions, enabling a potentially more efficient and effective way to search against a verifier, compared to na\"ive best-of-N~\citep{yao2023tree,feng2024alphazerolike,chen2024alphamath}.

\vspace{-0.25cm}
\section{How to Scale Test-Time Computation Optimally}
\label{sec:compute_optimal}
\vspace{-0.25cm}

Given the unification of various methods, we would now like to understand how to \emph{most effectively} utilize test-time computation to improve LM performance on a given prompt. Concretely we wish to answer:
\begin{AIbox}{Problem setup}
    We are given a prompt and a test-time compute budget within which to solve the problem. Under the abstraction above, there are different ways to utilize test-time computation. Each of these methods may be more or less effective depending on the specific problem given. How can we determine the \emph{most effective} way to utilize test-time compute for a given prompt? And how well would this do against simply utilizing a much bigger pretrained model?
\end{AIbox}
When either refining the proposal distribution or searching against a verifier, there are several different hyper-parameters that can be adjusted to determine how a test-time compute budget should be allocated. For example, when using a model finetuned for revisions as the proposal distribution and an ORM as the verifier, we could either spend the full test-time compute budget on generating N independent samples in parallel from the model and then apply best-of-N, or we could sample N revisions in sequence using a revision model and then select the best answer in the sequence with an ORM, or strike a balance between these extremes.
Intuitively, we might expect ``easier'' problems to benefit more from revisions, since the model's initial samples are more likely to be on the right track but may just need further refinement. On the other hand, challenging problems may require more exploration of different high-level problem solving strategies, so sampling many times independently in parallel may be preferable in this setting.

In the case of verifiers, we also have the option to choose between different search algorithms (e.g. beam-search, lookahead-search, best-of-N), each of which may exhibit different properties depending on the quality of the verifier and proposal distribution at hand. More sophisticated search procedures might be more useful in harder problems compared to a much simpler best-of-N or majority baseline.

\vspace{-0.3cm}
\subsection{Test-Time Compute-Optimal Scaling Strategy}
\vspace{-0.2cm}
In general, we would therefore like to select the \emph{optimal} allocation of our test-time compute budget for a given problem. To this end, for any given approach of utilizing test-time compute (e.g., revisions and search against a verifier in this paper, various other methods elsewhere), we define the \textbf{``test-time compute-optimal scaling strategy''} as the strategy that chooses hyperparameters corresponding to a given test-time strategy for maximal performance benefits on a given prompt at test time. Formally, define $\operatorname{Target}(\theta, N, q)$ as the distribution over natural language output tokens induced by the model for a given prompt $q$, using test-time compute hyper-parameters $\theta$, and a compute budget of $N$. We would like to select the hyper-parameters $\theta$ which maximize the accuracy of the target distribution for a given problem. We express this formally as:
\begin{align}
\label{eq:test_time_compute_optimal}
    \theta^{*}_{q,a^*(q)}(N) = \operatorname{argmax}_{\theta} \left( \mathbb{E}_{y \sim \operatorname{Target}(\theta, N, q)} \left[ \mathbbm{1}_{y = y^*(q)} \right] \right),
\end{align}
where $y^*(q)$ denotes the ground-truth correct response for $q$, and $\theta^{*}_{q,y^*(q)}(N)$ represents the test-time compute-optimal scaling strategy for the problem $q$ with compute budget $N$.

\vspace{-0.2cm}
\subsection{Estimating Question Difficulty for Compute-Optimal Scaling} 
\vspace{-0.2cm}
In order to effectively analyze the test-time scaling properties of the different mechanisms discussed in Section~\ref{sec:unifying_perspective} (e.g. the proposal distribution and the verifier), we will prescribe an approximation to this optimal strategy $\theta^{*}_{q,y^*(q)}(N)$ as a function of a statistic of a given prompt. This statistic estimates a notion of \emph{difficulty} for a given prompt. The compute-optimal strategy is defined as a function of the difficulty of this prompt. Despite being only an approximate solution to the problem shown in Equation~\ref{eq:test_time_compute_optimal}, we find that it can still induce substantial improvements in performance over a baseline strategy of allocating this inference-time compute in an ad-hoc or uniformly-sampled manner.

Our estimate of the question difficulty assigns a given question to one of five difficulty levels. We can then use this discrete difficulty categorization to estimate $\theta^{*}_{q,y^*(q)}(N)$ on a validation set for a given test-time compute budget. We then apply these compute-optimal strategies on the test-set. Concretely, we select the best performing test-time compute strategy for each difficulty bin independently. In this way, question difficulty acts as a sufficient statistic of a question when designing the compute-optimal strategy.

\textbf{Defining difficulty of a problem.} Following the approach of \citet{lightman2023lets}, we define question difficulty as a function of a given base LLM. Specifically, we bin the model's pass@1 rate -- estimated from 2048 samples -- on each question in the test set into five quantiles, each corresponding to increasing difficulty levels. We found this notion of model-specific difficulty bins to be more predictive of the efficacy of using test-time compute in contrast to the hand-labeled difficulty bins in the MATH dataset.

That said, we do note that assessing a question's difficulty as described above assumes oracle access to a ground-truth correctness checking function, which is of course not available upon deployment where we are only given access to test prompts that we don't know the answer to. In order to be feasible in practice, a compute-optimal scaling strategy conditioned on difficulty needs to first assess difficulty and then utilize the right scaling strategy to solve this problem.
Therefore, we approximate the problem's difficulty via a \textbf{model-predicted notion of difficulty}, which performs the same binning procedure over the the averaged final answer score from a learned verifier (and not groundtruth answer correctness checks) on the same set of 2048 samples per problem. We refer to this setting as \textbf{model-predicted difficulty} and the setting which relies on the ground-truth correctness as \textbf{oracle difficulty}.

While model-predicted difficulty removes the need for need knowing the ground truth label, estimating difficulty in this way still incurs additional computation cost during inference. That said, this one-time inference cost can be subsumed within the cost for actually running an inference-time strategy (e.g., when using a verifier, one could use the same inference computation for also running search). More generally, this is akin to exploration-exploitation tradeoff in reinforcement learning: in actual deployment conditions, we must balance the compute spent in assessing difficulty vs applying the most compute-optimal approach. This is a crucial avenue for future work (see Section~\ref{sec:discussion}) and our experiments do not account for this cost largely for simplicity, since our goal is to present some of the first results of \emph{what is in fact possible} by effectively allocating test-time compute.

So as to avoid confounders with using the same test set for computing difficulty bins and for selecting the compute-optimal strategy, we use two-fold cross validation on each difficulty bin in the test set. We select the best-performing strategy according to performance on one fold and then measure performance using that strategy on the other fold and vice versa, averaging the results of the two test folds.

\vspace{-0.2cm}
\section{Experimental Setup}
\label{sec:setup}
\vspace{-0.2cm}
We first outline our experimental setup for conducting this analysis with multiple verifier design choices and proposal distributions, followed by the analysis results in the subsequent sections.

\textbf{Datasets.} We expect test-time compute to be most helpful when models already have all the basic ``knowledge'' needed to answer a question, and instead the primary challenge is about drawing (complex) inferences from this knowledge. To this end, we focus on the MATH~\citep{hendrycks2021measuring} benchmark, which consists of high-school competition level math problems with a range of difficulty levels. For all experiments, we use the dataset split consisting of 12k train and 500 test questions, used in~\citet{lightman2023lets}.

\textbf{Models.} We conduct our analysis using the PaLM 2-S*~\citep{anil2023palm} (Codey) base model. We believe this model is representative of the capabilities of many contemporary LLMs, and therefore think that our findings likely transfer to similar models. Most importantly, this model attains a non-trivial performance on MATH and yet has not saturated, so we expect this model to provide a good test-bed for us.

\vspace{-0.2cm}
\section{Scaling Test-Time Compute via Verifiers}
\label{sec:search}
\vspace{-0.2cm}
In this section we analyze how test-time compute can be scaled by optimizing a verifier, as effectively as possible. To this end, we study different approaches for performing test-time search with process verifiers (PRMs) and analyze the test-time compute scaling properties of these different approaches.

\vspace{-0.2cm}
\subsection{Training Verifiers Amenable to Search}
\vspace{-0.1cm}

\textbf{PRM training.} Originally PRM training~\citep{uesato2022solving,lightman2023lets} used human crowd-worker labels.
While~\citet{lightman2023lets} released their PRM training data (i.e., the PRM800k dataset), we found this data to be largely ineffective for us. We found that it was easy to exploit a PRM trained on this dataset via even na\"ive strategies such as best-of-N sampling. We hypothesize that this is likely a result of the distribution shift between the GPT-4 generated samples in their dataset and our PaLM 2 models.
Rather than proceeding with the expensive process of collecting crowd-worker PRM labels for our PaLM 2 models, we instead apply the approach of~\citet{wang2023mathshepherd} to supervise PRMs without human labels, using estimates of per-step correctness obtained from running Monte Carlo rollouts from each step in the solution.
Our PRM's per-step predictions therefore correspond to value estimates of reward-to-go for the base model's sampling policy, similar to recent work~\citep{wang2023mathshepherd,setlur2024rl}. We also compared to an ORM baseline (Appendix~\ref{app:comparing_prm_orm}) but found that our PRM consistently outperforms the ORM.
Hence, all of the search experiments in this section use a PRM model. Additional details on PRM training are shown in Appendix~\ref{app:prm_training}.

\textbf{Answer aggregation.}
At test time, process-based verifiers can be used to score each individual step in a set of solutions sampled from the base model. In order to select the best-of-N answers with the PRM, we need a function that can aggregate across all the per-step scores for each answer to determine the best candidate for the correct answer. To do this, we first aggregate each individual answer's per-step scores to obtain a final score for the full answer (step-wise aggregation). We then aggregate across answers to determine the best answer (inter-answer aggregation). Concretely, we handle step-wise and inter-answer aggregation as follows:
\begin{itemize}
    \vspace{-0.2cm}
    \item \textbf{Step-wise aggregation.} Rather than aggregating the per-step scores by taking the product or minimum~\citep{wang2023mathshepherd,lightman2023lets}, we instead use the PRM's prediction at the last step as the full-answer score. We found this to perform the best out of all aggregation methods we studied (see  Appendix~\ref{app:comparing_prm_agg}).
    \item \textbf{Inter-answer aggregation.} We follow~\citet{li2023making} and apply ``best-of-N weighted'' selection rather than standard best-of-N. Best-of-N weighted selection marginalizes the verifier's correctness scores across all solutions with the same final answer, selecting final answer with the greatest total sum.
\end{itemize}

\vspace{-0.4cm}
\subsection{Search Methods Against a PRM}
\vspace{-0.15cm}
\begin{figure}[t]
    \centering
    \vspace{-0.2cm}
    \includegraphics[width=0.95\linewidth]{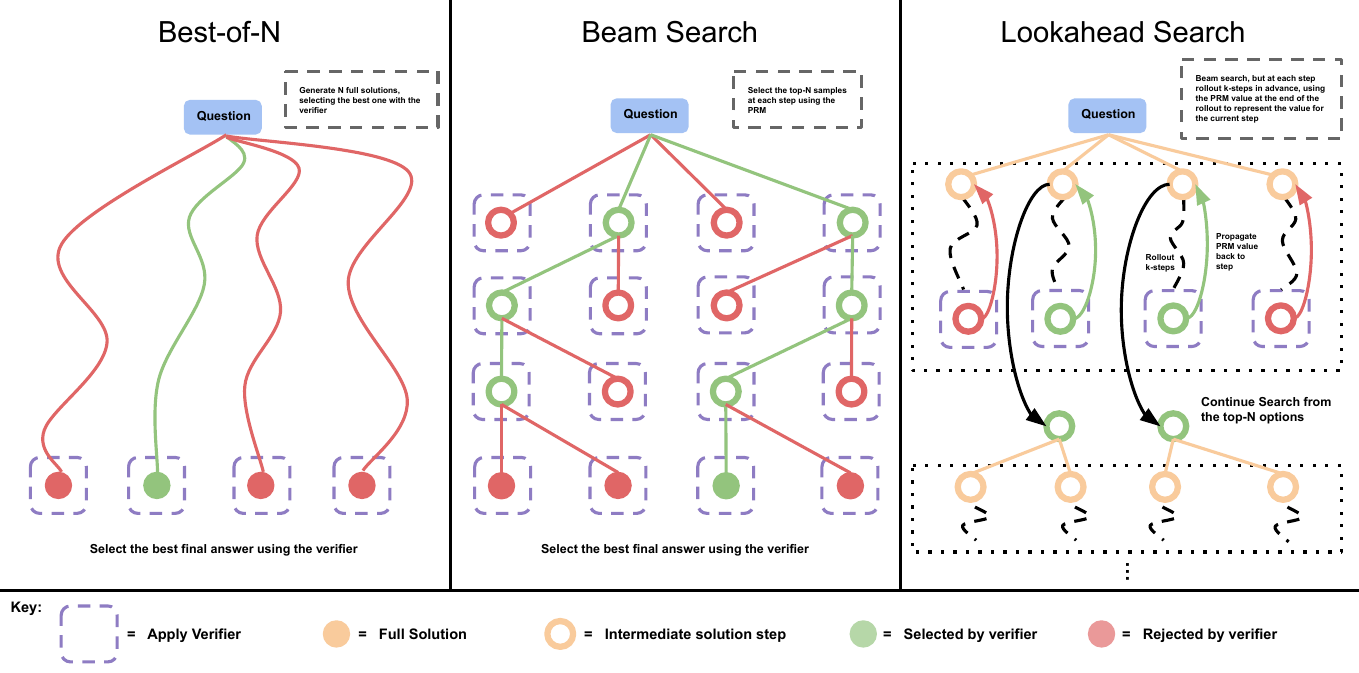}
    \vspace{-0.25cm}
    \caption{\footnotesize{\textbf{\emph{Comparing different PRM search methods.}} \textbf{Left:} Best-of-N samples N full answers and then selects the best answer according to the PRM final score. \textbf{Center:} Beam search samples N candidates at each step, and selects the top M according to the PRM to continue the search from. \textbf{Right:} lookahead-search extends each step in beam-search to utilize a k-step lookahead while assessing which steps to retain and continue the search from. Thus lookahead-search needs more compute.}}
    \label{fig:search_methods}
    \vspace{-0.2cm}
\end{figure}
We optimize the PRM at test time via search methods. We study three search approaches that sample outputs from a few-shot prompted base LLM (see Appendix~\ref{app:prompting}). An illustration is shown in Figure~\ref{fig:search_methods}.

\textbf{Best-of-N weighted.} We sample N answers independently from the base LLM and then select the best answer according to the PRM's final answer judgement.

\textbf{Beam search.} Beam search optimizes the PRM by searching over its per-step predictions. Our implementation is similar to BFS-V~\citep{yao2023tree,feng2024alphazerolike}. Concretely, we consider a fixed number of beams $N$ and a beam width $M$. We then run the following steps:
\begin{enumerate}
\vspace{-0.2cm}
    \item sample $N$ initial predictions for the first step in the solution
    \item score the generated steps according to the PRM's predicted step-wise reward-to-go estimate (which also corresponds to the total reward from the prefix since the reward is sparse in this setting)
    \item filter for only the top $\frac{N}{M}$ highest scoring steps
    \item now from each candidate, sample $M$ proposals from the next step, resulting in a total of $N/M \times M$ candidate prefixes again. Then repeat steps 2-4 again. \vspace{-0.3cm}
\end{enumerate}
We run this algorithm until the end of a solution or the maximum number of rounds of beam expansion are attained (40 in our case). We conclude the search with N final answer candidates, to which we apply best-of-N weighted selection described above to make our final answer prediction.

\textbf{Lookahead search.} Lookahead search modifies how beam search evaluates individual steps.
It uses lookahead rollouts to improve the accuracy of the PRM's value estimation in each step of the search process.
Specifically, at each step in the beam search, rather than using the PRM score at the current step to select the top candidates, lookahead search performs a simulation, rolling out up to $k$ steps further while stopping early if the end of solution is reached. To minimize variance in the simulation rollout, we perform rollouts using temperature 0.
The PRM's prediction at the end of this rollout is then used to score the current step in the beam search.
That is, in other words, we can view beam search as a special case of lookahead search with $k=0$.
Given an accurate PRM, increasing $k$ should improve the accuracy of the per-step value estimates at the cost of additional compute.
Also note that this version of lookahead search is a special case of MCTS~\citep{suttonrlbook}, wherein the stochastic elements of MCTS, designed to facilitate exploration, are removed since the PRM is already trained and is frozen.
These stochastic elements are largely useful for learning the value function (which we've already learned with our PRM), but less useful at test-time when we want to exploit rather than explore.
Therefore, lookahead search is largely representative of how MCTS-style methods would be applied at test-time.

\begin{figure}
    \centering
    \includegraphics[width=0.99\textwidth]{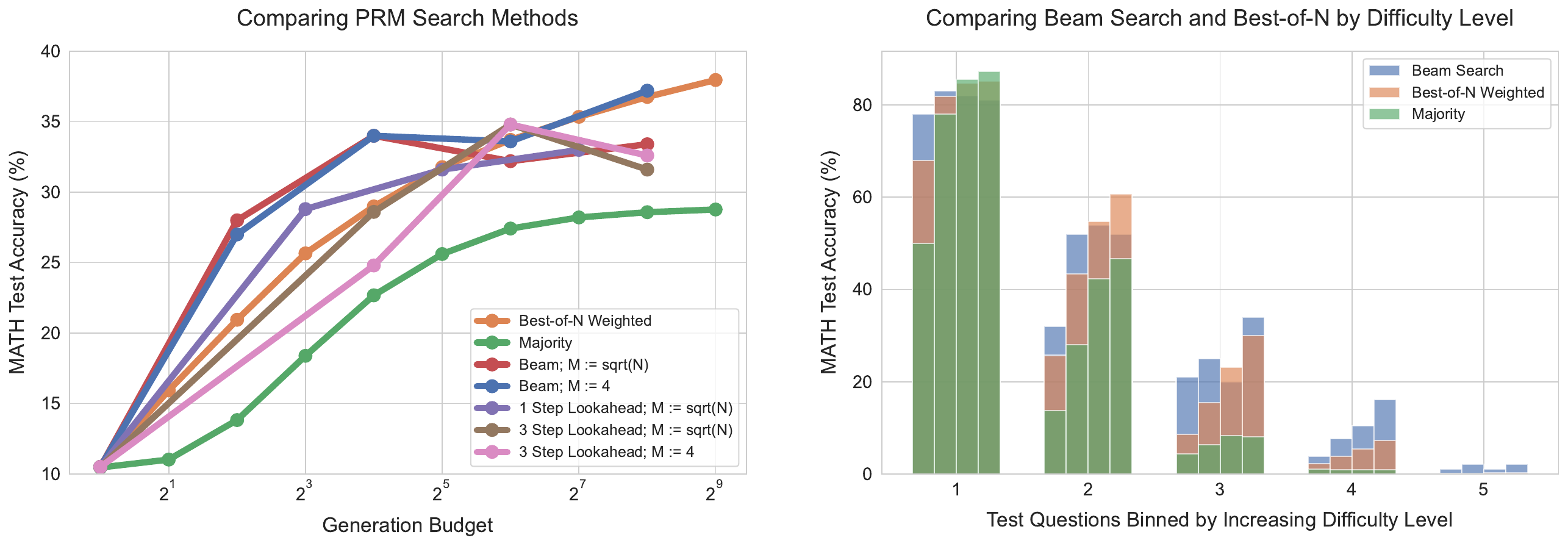}
    \vspace{-0.25cm}
    \caption{\footnotesize{\textbf{Left:} \textbf{\emph{Comparing different methods for conducting search against PRM verifiers}}. We see that at low generation budgets, beam search performs best, but as we scale the budget further the improvements diminish, falling below the best-of-N baseline. Lookahead-search generally underperforms other methods at the same generation budget. \textbf{Right:} \textbf{\emph{Comparing beam search and best-of-N binned by difficulty level}}. The four bars in each difficulty bin correspond to increasing test-time compute budgets (4, 16, 64, and 256 generations). On the easier problems (bins 1 and 2), beam search shows signs of over-optimization with higher budgets, whereas best-of-N does not. On the medium difficulty problems (bins 3 and 4), we see beam search demonstrating consistent improvements over best-of-N.}}
    \label{fig:comparing_search_and_beam_difficulty}
    \vspace{-0.2cm}
\end{figure}

\vspace{-0.2cm}
\subsection{Analysis Results: Test-Time Scaling for Search with Verifiers}
\vspace{-0.2cm}
We now present our results comparing various search algorithms and identify a prompt difficulty dependent compute-optimal scaling strategy for search methods.

\textbf{Comparing search algorithms.} We first conduct a sweep over various search settings. In addition to the standard best-of-N approach, we sweep over the two main parameters that distinguish different tree-search methods: beam-width $M$ and number of lookahead steps $k$. While we are not able to extensively sweep every single configuration, we sweep over the following settings with a maximum budget of 256:

\renewcommand\labelenumi{\theenumi)}
\begin{enumerate}
\vspace{-0.35cm}
    \item Beam search with the beam width set to $\sqrt{N}$, where $N$ is the generation budget.
    \item Beam search with a fixed beam width of 4.
    \item Lookahead search with $k=3$ applied to both beam-search settings 1) and 2).
    \item Lookahead search with $k=1$ applied to beam-search setting 1). \vspace{-0.3cm}
\end{enumerate}
To compare search methods as a function of generation budget fairly, we build a protocol for estimating the cost of each method. We consider a generation to be a sampled answer from the base LLM. For beam search and best-of-N the generation budget corresponds to the number of beams and $N$ respectively. Lookahead search, however, utilizes additional compute: at each step of the search, we sample $k$ additional steps ahead. Therefore, we define the cost of lookahead-search to be $N \times (k+1)$ samples. 

\textbf{Results.} As shown in Figure~\ref{fig:comparing_search_and_beam_difficulty} (left), with smaller generation budgets, beam search significantly outperforms best-of-N. However, as the budget is scaled up, these improvements greatly diminish, with beam search often underperforming the best-of-N baseline. We also see that, lookahead-search generally underperforms other methods at the same generation budget, likely due to the additional computation inducted by simulating the lookahead rollouts.
The diminishing returns from search are likely due to exploitation of the PRM's predictions. For example, we see some instances (such as in Figure~\ref{fig:prm_ex6}), where search causes the model to generate low-information repetitive steps at the end of a solution. In other cases, we find that over-optimizing search can result in overly short solutions consisting of just 1-2 steps.
This explains why the most powerful search method (i.e., lookahead search) underperforms the most. We include several of these examples found by search in Appendix~\ref{app:prm_example_outputs}.

\textbf{Which problems does search improve?} To understand how to compute-optimally scale search methods, we now conduct a difficulty bin analysis. Specifically, we compare beam-search ($M=4$) against best-of-N. In Figure~\ref{fig:comparing_search_and_beam_difficulty} (right) we see that while in aggregate, beam search and best-of-N perform similarly with a high generation budget, evaluating their efficacy over difficulty bins reveals very different trends.
On the easy questions (levels 1 and 2), the stronger optimizer of the two approaches, beam search, degrades performance as the generation budget increases, suggesting signs of exploitation of the PRM signal.
In contrast, on the harder questions (levels 3 and 4), beam search consistently outperforms best-of-N. On the most difficult questions (level 5), no method makes much meaningful progress. 

These findings match intuition: we might expect that on the easy questions, the verifier will make mostly correct assessments of correctness.  Therefore, by applying further optimization via beam search, we only further amplify any spurious features learned by the verifier, causing performance degredation.
On the more difficult questions, the base model is much less likely to sample the correct answer in the first place, so search can serve to help guide the model towards producing the correct answer more often.

\begin{wrapfigure}{r}{0.5\textwidth}
    \vspace{-0.5cm}
  \centering
  \includegraphics[width=0.48\textwidth]{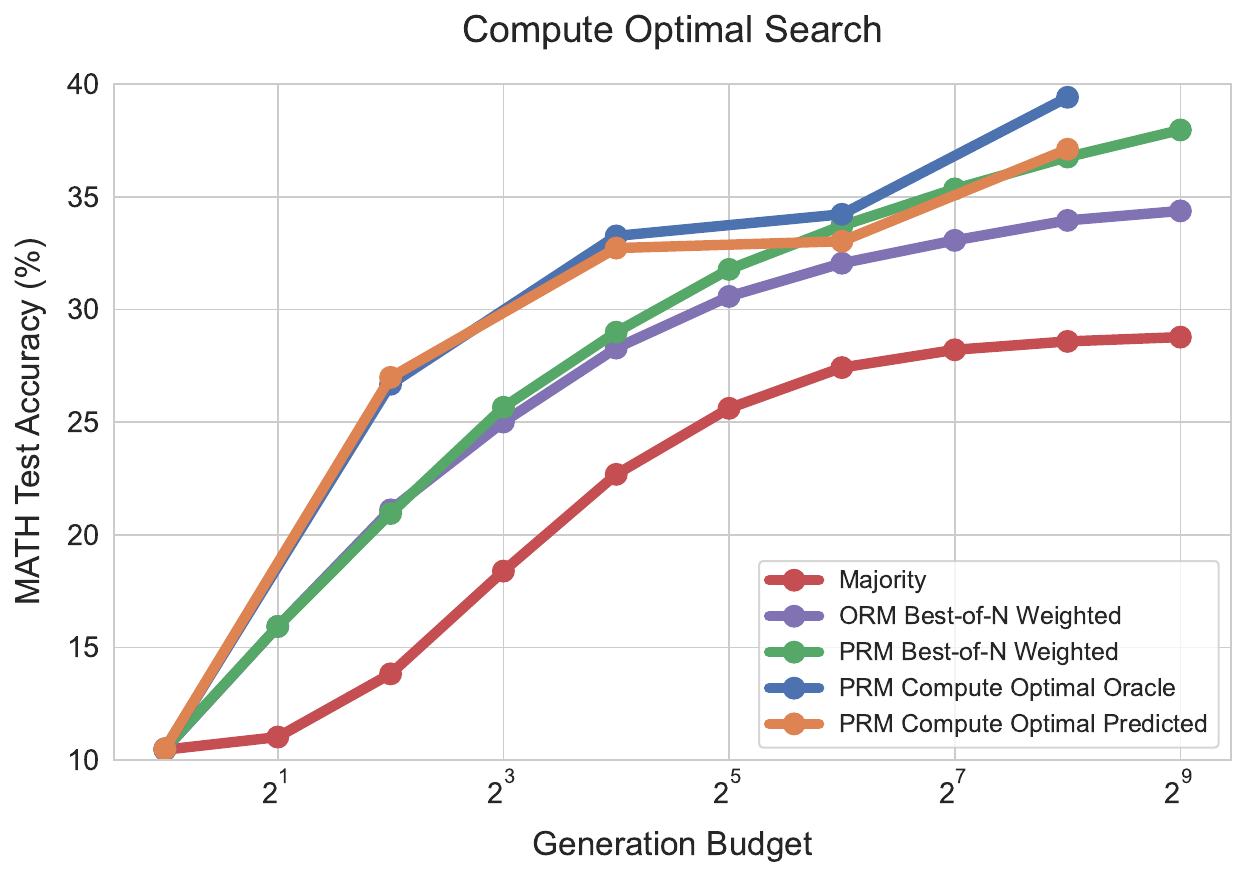}
  \vspace{-0.3cm}
  \caption{\footnotesize{\textbf{\emph{Comparing compute-optimal test-time compute allocation against baselines with PRM search.}} By scaling test time compute per the notion of question difficulty, we find that we can nearly outperform PRM best-of-N using up to \textbf{4$\times$} less test-time compute (e.g. 16 verses 64 generations). ``\textbf{Compute-optimal oracle}'' refers to using oracle difficulty bins derived from the groundtruth correctness information, and ``\textbf{compute-optimal predicted}'' refers to using the PRM's predictions to generate difficulty bins. Observe that the curves with either type of difficulty bins largely overlap with each other.}}
  \label{fig:compute_optimal_search}
  \vspace{-0.4cm}
\end{wrapfigure}
\textbf{Compute-optimal search.} Given the above results, it is clear that question difficulty can be a useful statistic to predict the optimal search strategy to use at a given compute budget. 
Additionally, the best choice of search strategy can vary drastically as a function of this difficulty statistic. We therefore visualize the ``compute-optimal'' scaling trend, as represented by the best performing search strategy at each difficulty level in Figure~\ref{fig:compute_optimal_search}.
We see that in the low generation budget regime, using both the oracle and predicted difficulty, \textbf{compute-optimal scaling can nearly outperform best-of-N using up to} \emph{\textbf{4x}} \textbf{less test-time compute} (e.g. 16 verses 64 generations). While in the higher budget regime, some of these benefits diminish with the use of predicted difficulty, with oracle bins we still see continued improvements from optimally scaling test-time compute. This result demonstrates the performance gains that could be obtained by adaptively allocating test-time compute during search.

\begin{AIbox}{Takeaways for compute-optimal scaling of verifiers}
We find that the efficacy of any given verifier search method depends critically on both the compute budget and the question at hand. Specifically, beam-search is more effective on harder questions and at lower compute budgets, whereas best-of-N is more effective on easier questions and at higher budgets. Moreover, by selecting the best search setting for a given question difficulty and test-time compute budget, we can nearly outperform best-of-N using up to \emph{4x} less test-time compute.
\end{AIbox}

\vspace{-0.2cm}
\section{Refining the Proposal Distribution}
\label{sec:revisions}
\vspace{-0.2cm}
So far, we studied the test-time compute scaling properties of search against verifiers. Now we turn to studying the scaling properties of modifying the proposal distribution (Section~\ref{sec:unifying_perspective}).
Concretely, we enable the model to revise their own answers iteratively, allowing the model to dynamically improve it's own distribution at test time. Simply prompting existing LLMs to correct their own mistakes tends to be largely ineffective for obtaining performance improvements on reasoning problems~\citep{huang2023large}. Therefore, we build on the recipe prescribed by \citet{qu2024recursive}, incorporate modifications for our setting, and finetune language models to iteratively revise their own answers. We first describe how we train and use models that refine their own proposal distribution by sequentially conditioning on their own previous attempts at the question. We then analyze the inference-time scaling properties of revision models.

\begin{figure}
    \centering
    \includegraphics[width=0.99\linewidth]{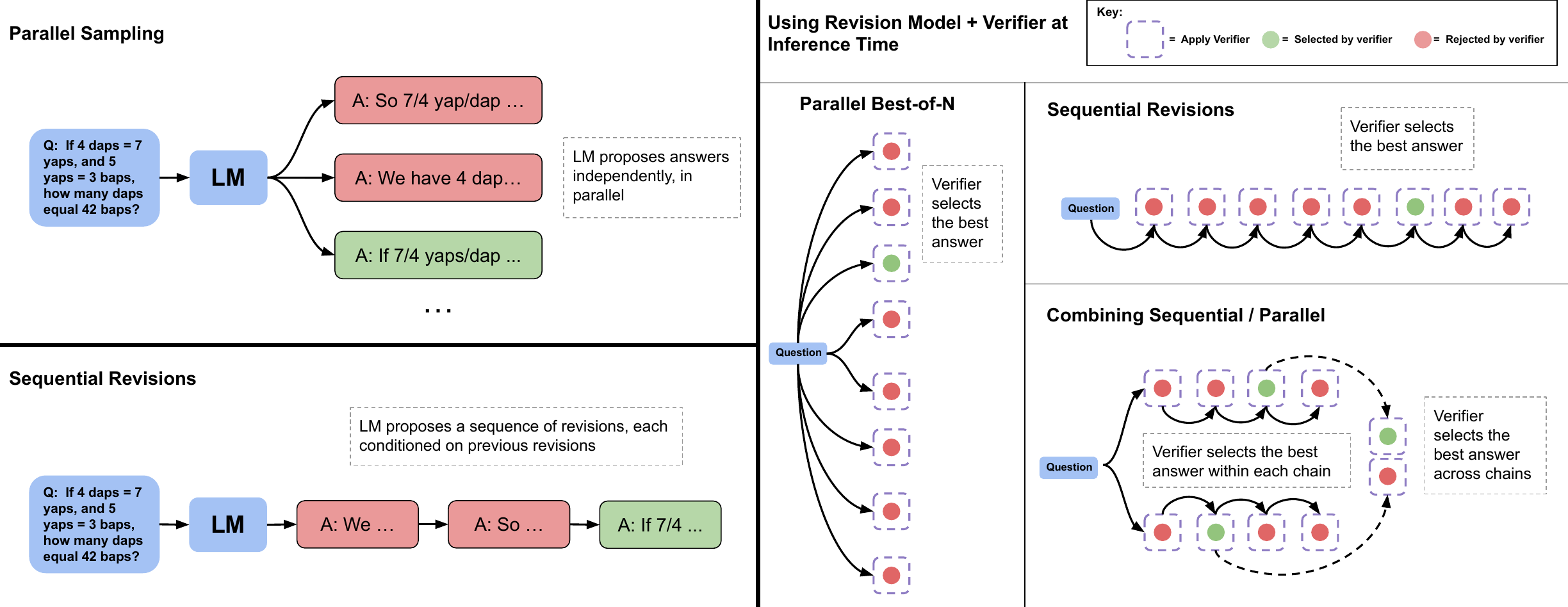}
    \vspace{-0.3cm}
    \caption{\footnotesize{\textbf{\emph{Parallel sampling (e.g., Best-of-N) verses sequential revisions.}} \textbf{Left:} Parallel sampling generates N answers independently in parallel, whereas sequential revisions generates each one in sequence conditioned on previous attempts. \textbf{Right:} In both the sequential and parallel cases, we can use the verifier to determine the best-of-N answers (e.g. by applying best-of-N weighted). We can also allocate some of our budget to parallel and some to sequential, effectively enabling a combination of the two sampling strategies. In this case, we use the verifier to first select the best answer within each sequential chain and then select the best answer accross chains.}}
    \label{fig:how_revisions_works}
    \vspace{-0.3cm}
\end{figure}

\vspace{-0.3cm}
\subsection{Setup: Training and Using Revision Models}
\label{sec:revision_setup}
\vspace{-0.2cm}
Our procedure for finetuning revision models is similar to~\citep{qu2024recursive}, though we introduce some crucial differences.
For finetuning, we need trajectories consisting of a sequence of incorrect answers followed by a correct answer, that we can then run SFT on. Ideally, we want the correct answer to be \emph{correlated} with the incorrect answers provided in context, so as to effectively teach the model to \emph{implicitly} identify mistakes in examples provided in-context, followed by correcting those mistakes by making edits as opposed to ignoring the in-context examples altogether, and trying again from scratch.

\textbf{Generating revision data.} The on-policy approach of \citet{qu2024recursive} for obtaining several multi-turn rollouts was shown to be effective, but it was not entirely feasible in our infrastructure due to compute costs associated with running multi-turn rollouts. Therefore, we sampled 64 responses \emph{in parallel} at a higher temperature and post-hoc constructed multi-turn rollouts from these independent samples. Specifically, following the recipe of \citep{anonymousrevisions}, we pair up each correct answer  with a sequence of incorrect answers from this set as context to construct multi-turn finetuning data.
We include up to four incorrect answers in context, where the specific number of solutions in context is sampled randomly from a uniform distribution over categories 0 to 4. We use a character edit distance metric to prioritize selecting incorrect answers which are correlated with the final correct answer (see Appendix~\ref{app:revision_finetune}). Note that token edit distance is not a perfect measure of correlation, but we found this heuristic to be sufficient to correlate incorrect in-context answers with correct target answers to facilitate training a meaningful revision model, as opposed to randomly pairing incorrect and correct responses with uncorrelated responses.

\begin{figure}
    \centering
    \includegraphics[width=0.99\textwidth]{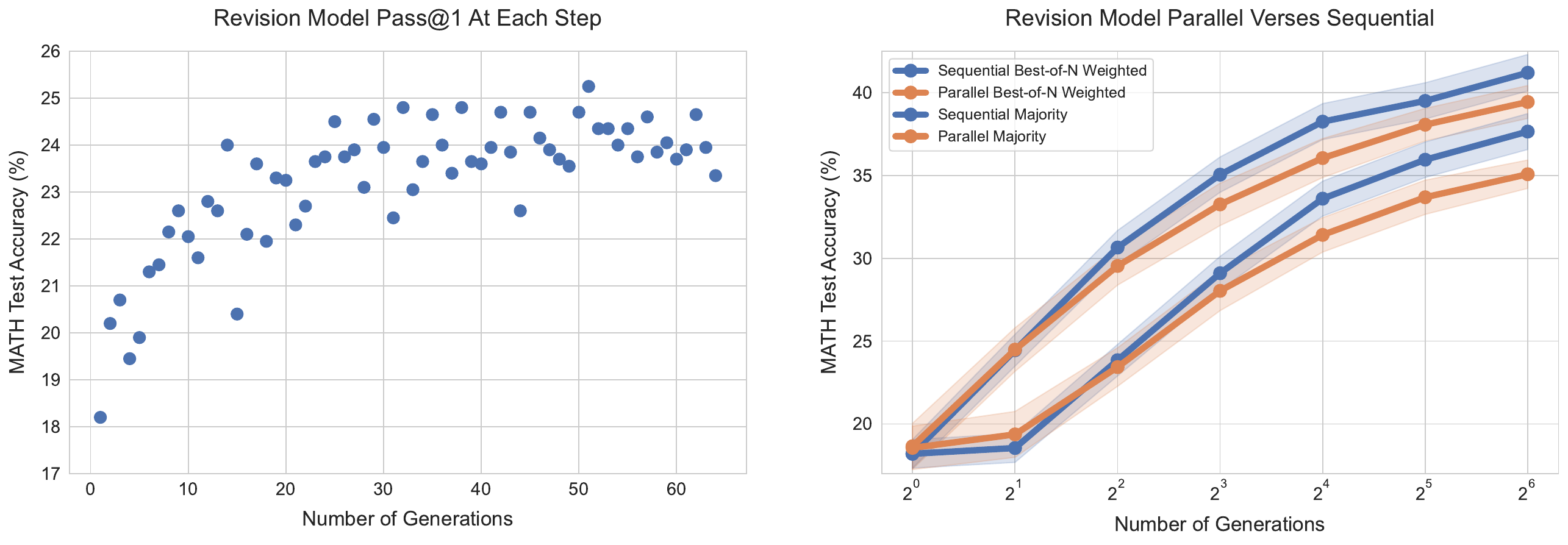}
    \vspace{-0.3cm}
    \caption{\footnotesize{\textbf{Left:} \textbf{\emph{Our revision model's pass@1 at each revision step.}} Pass@1 gradually improves after each revision step, even improving beyond the 4 revision steps that it was trained for. We estimate pass@1 at each step by by averaging over the performance of 4 revision trajectories of length 64 for each question in the test-set. \textbf{Right:} \textbf{\emph{Sequential vs parallel sampling from the revision model.}} Comparing performance when generating N initial answers in parallel from our revision model, verses generating N revisions sequentially, with the model. When using both the verifier and majority voting to select the answer, we see that generating answers sequentially with the revision model narrowly outperforms generating them in parallel.}}
    \label{fig:revision_model_results}
    \vspace{-0.3cm}
\end{figure}

\textbf{Using revisions at inference-time.} Given a finetuned revision model, we can then sample a sequence of revisions from the model at test-time. While our revision model is only trained with up to four previous answers in-context, we can sample longer chains by truncating the context to the most recent four revised responses. In Figure~\ref{fig:revision_model_results} (left), we see that as we sample longer chains from the revision model, the model's pass@1 at each step gradually improves, demonstrating that we are able to effectively teach the model to learn from mistakes made by previous answers in context.

That said, there is a distribution shift at inference time: the model was trained on only sequences with incorrect answers in context, but at test-time the model may sample correct answers that are included in the context.
In this case, it may incidentally turn the correct answer into an incorrect answer in the next revision step.
We find that indeed, similar to~\citet{qu2024recursive}, around 38\% of correct answers get converted back to incorrect ones with our revision  model using a na\"ive approach. Therefore, we employ a mechanism based on sequential majority voting or verifier-based selection to select the most correct answer from the sequence of revisions made by the model (see Figure~\ref{fig:how_revisions_works}) to produce the best answer.

\textbf{Comparisons.} To test the efficacy of modifying the proposal distribution via revisions, we setup an even comparison between the performance of sampling N revisions in sequence and sampling N attempts at a question in parallel. We see in Figure~\ref{fig:revision_model_results} (right), that with both the verifier-based and majority-based selection mechanisms sampling solutions in sequence outperforms sampling them in parallel.

\begin{figure}
    \centering
    \includegraphics[width=0.99\textwidth]{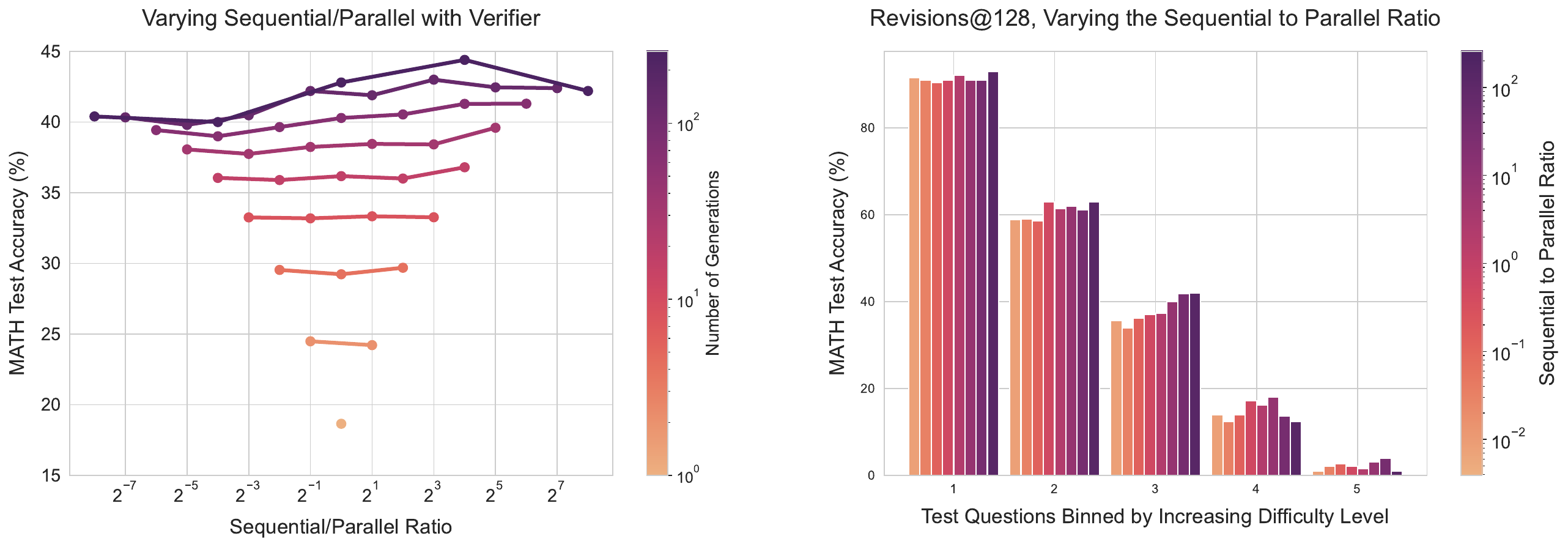}
    \vspace{-0.3cm}
    \caption{\footnotesize{\textbf{Left:} \emph{\textbf{Varying the ratio of the generation budget allocated sequential revisions to verses parallel samples.}} Each line represents a fixed generation budget as the ratio is changed. We use the verifier for answer selection. We see that while increased sequential revisions tends to outperform more parallel compute, at higher generation budgets there is an ideal ratio that strikes a balance between the two extremes. \textbf{Right:} \textbf{\emph{Varying the sequential to parallel ratio for a generation budget of 128 across difficulty bins.}} Using verifier-based selection, we see that the easier questions attain the best performance with full sequential compute. On the harder questions, there is an ideal ratio of sequential to parallel test-time compute.}}
    \label{fig:iso_revisions}
    \vspace{-0.25cm}
\end{figure}
 
\vspace{-0.2cm}
\subsection{Analysis Results: Test-Time Scaling with Revisions}
\vspace{-0.2cm}

We saw previously that proposing answers sequentially outperforms proposing them in parallel. However, we might expect sequential and parallel sampling to have different properties.
Sampling answers in parallel may act as more of a global search process, that could in principle, provide coverage over many totally different approaches for solving a problem, for instance, different candidates might utilize different high-level approaches altogether.
Sequential sampling, on the other hand, may work more as a local refinement process, revising responses that are already somewhat on the right track.
Due to these complementary benefits, we should strike a balance between these two extremes by allocating some of our inference-time budget to parallel sampling (e.g. $\sqrt{N}$) and the rest to sequential revisions (e.g. $\sqrt{N}$). We will now show the existence of a compute-optimal ratio between sequential and parallel sampling, and understand their relative pros and cons based on difficulty of a given prompt.

\textbf{Trading off sequential and parallel test-time compute.} To understand how to optimally allocate sequential and parallel compute, we perform a sweep over a number of different ratios. We see, in Figure~\ref{fig:iso_revisions} (left), that indeed, at a given generation budget, \textbf{there exists an ideal sequential to parallel ratio, that achieves the maximum accuracy.} We also see in Figure~\ref{fig:iso_revisions} (right) that {the ideal ratio of sequential to parallel varies depending on a given question's difficulty.} In particular, easy questions benefit more from sequential revisions, whereas on difficult questions it is optimal to strike a balance between sequential and parallel computation. This finding supports the hypothesis that sequential revisions (i.e., varying the proposal distribution) and parallel sampling (i.e., search with verifiers) are two complementary axes for scaling up test-time compute, which may be more effective on a per-prompt basis. We include examples of our model's generations in Appendix~\ref{app:revision_example_outputs}. Additional results are shown in Appendix~\ref{app:additional_revision}.

\begin{wrapfigure}{r}{0.5\textwidth}
  \centering
  \vspace{-0.5cm}
  \includegraphics[width=0.48\textwidth]{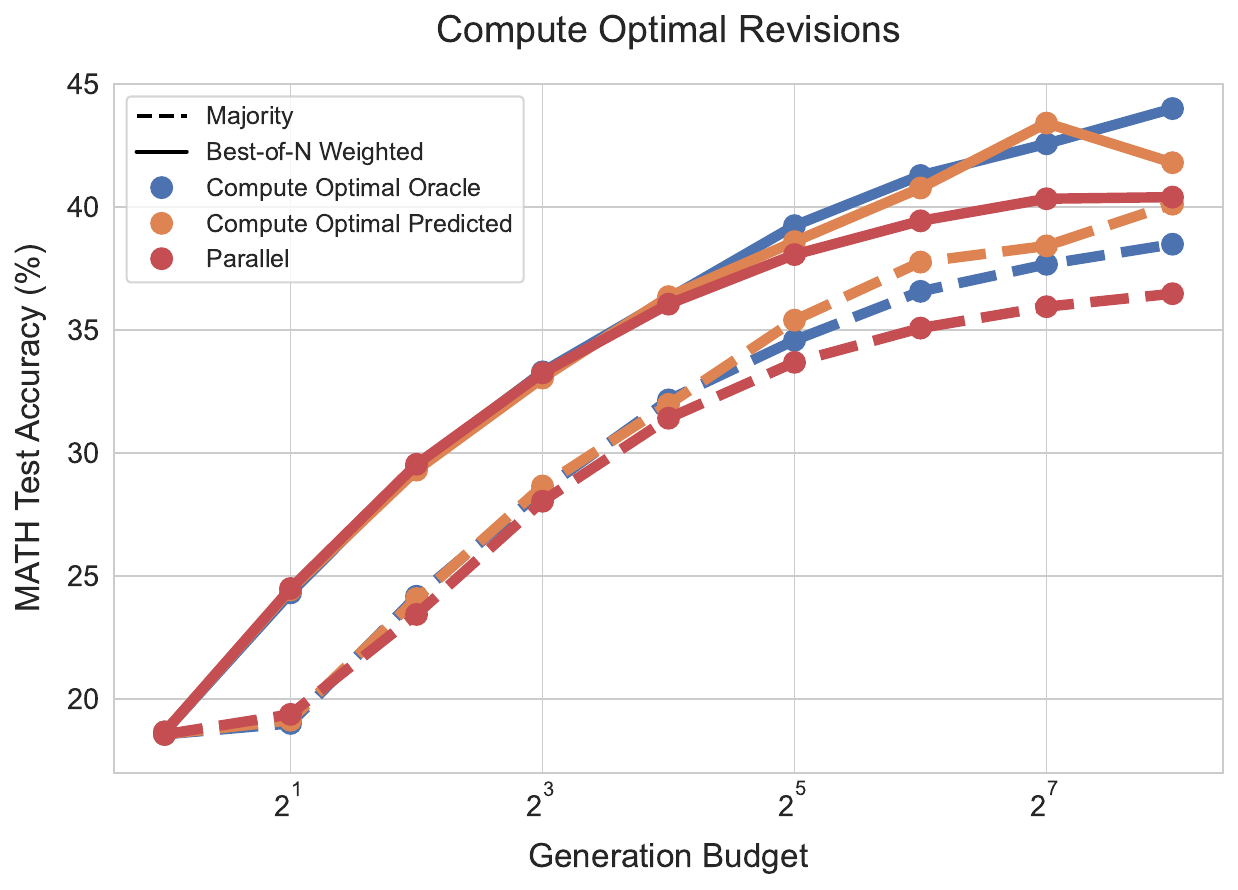}
  \vspace{-0.25cm}
  \caption{\footnotesize{\textbf{\emph{Comparing compute-optimal test-time compute allocation against the parallel compute baseline with our revision model}}. By optimally scaling test-time compute according to question difficulty, we find that we can outperform best-of-N using up to \emph{\textbf{4x}} less test-time compute (e.g. 64 samples verses 256). ``\textbf{Compute-optimal oracle}'' refers to using the oracle difficulty bins derived from the ground truth correctness information, and ``\textbf{compute optimal predicted}'' refers to using the PRM's predictions to produce model-predicted difficulty bins.}}
  \label{fig:compute_optimal_revisions}
  \vspace{-0.7cm}
\end{wrapfigure}
\textbf{Compute-optimal revisions.} Given that the efficacy of sequential and parallel sampling depends on question difficulty, we can select the ideal ratio of sequential to parallel compute per difficulty bin. In Figure~\ref{fig:compute_optimal_revisions}, we plot results using this compute-optimal scaling strategy when employing both our oracle and predicted notions of difficulty. In both cases, we're able to substantially improve test-time compute scaling by improving the proposal distribution via revisions. In particular, we see that at higher generation budgets, parallel sampling seems to plateau, whereas compute-optimal scaling demonstrates continued improvements. For both oracle and predicted difficulty bins, we see that \textbf{compute-optimal scaling can outperform best-of-N using up to} \emph{\textbf{4x}} \textbf{less test-time compute} (e.g. 64 samples verses 256). Overall, these results demonstrate the potential for improved test-time compute scaling by adjusting the proposal distribution on a per-prompt basis.

\begin{AIbox}{Takeaways for compute-optimal scaling by refining the proposal distribution with revisions}
We find that there exists a tradeoff between sequential (e.g. revisions) and parallel (e.g. standard best-of-N) test-time computation, and the ideal ratio of sequential to parallel test-time compute depends critially on both the compute budget and the specific question at hand. Specifically, easier questions benefit from purely sequential test-time compute, whereas harder questions often perform best with some ideal ratio of sequential to parallel compute. Moreover, by optimally selecting the best setting for a given question difficulty and test-time compute budget, we can outperform the parallel best-of-N baseline using up to \emph{4x} less test-time compute.
\end{AIbox}

\vspace{-0.25cm}
\section{Putting it Together: Exchanging Pretraining and Test-Time Compute}
\label{sec:exchanging}
\vspace{-0.2cm}
So far, we saw that utilizing additional test-time computation can enable us to represent more complex distributions than the one predicted by the base LLM itself, thereby improving performance. We now posit that this increased flexibility of representing distributions means that we can expect additional test-time compute to make up for the lack of a higher-capacity model or training for more FLOPs during pre-training. In this section, \emph{\textbf{we study to what extent this is possible.}} We pose the following question:

\begin{AIbox}{Question: Exchanging pretraining and test-time compute}
    Suppose a model was pre-trained with $X$ FLOPs. Assume that we plan to run $Y$ FLOPs of inference with this model. If we want to improve performance by increasing the total FLOPs budget by a factor of $M$ (i.e., $M(X+Y)$ total FLOPs across both pretraining and inference), should we spend our FLOPs on increased pretraining compute or on additional test-time compute?
\end{AIbox}
Increasing pretraining FLOPS introduces the additional design decision of whether to allocate compute to training with more data or more parameters~\citep{hoffmann2022training}. We focus on the setting in which model parameters are scaled up and training data amount is fixed, matching the approach taken with the open-source LLaMA series of models~\citep{touvron2023llama2openfoundation}. We choose this setting as it is representative of a canonical approach to scaling pretraining compute and leave the analysis of compute-optimal scaling of pretraining compute~\citep{sardana2023chinchillaoptimal} where the data and parameters are both scaled equally to future work.

\textbf{Defining an exchange rate between FLOPs.} We now describe how we define the exchange rate between pretraining and inference FLOPs. To determine pretraining FLOPs, use use the common approximation $X = 6ND_{\text{pretrain}}$~\citep{hoffmann2022training}, and for inference FLOPs, we use $Y = 2ND_{\text{inference}}$~\citep{sardana2023chinchillaoptimal}. Here $N$ represents model parameters, $D_{\text{pretrain}}$ is the number of tokens used for pretraining, and $D_{\text{inference}}$ the total number of tokens generated at inference time. With these approximations, we can see that, if we multiply the model parameters by a factor of $M$, then both the pretraining and inference FLOPs (due to the cost of greedy decoding with the larger model), increase by a factor of $M$ (giving $M(X+Y)$ total FLOPs).

\begin{figure}[t]
    \centering
    \vspace{-0.3cm}
    \includegraphics[trim=2.5cm 0cm 2.5cm 0cm, clip, width=0.99\textwidth]{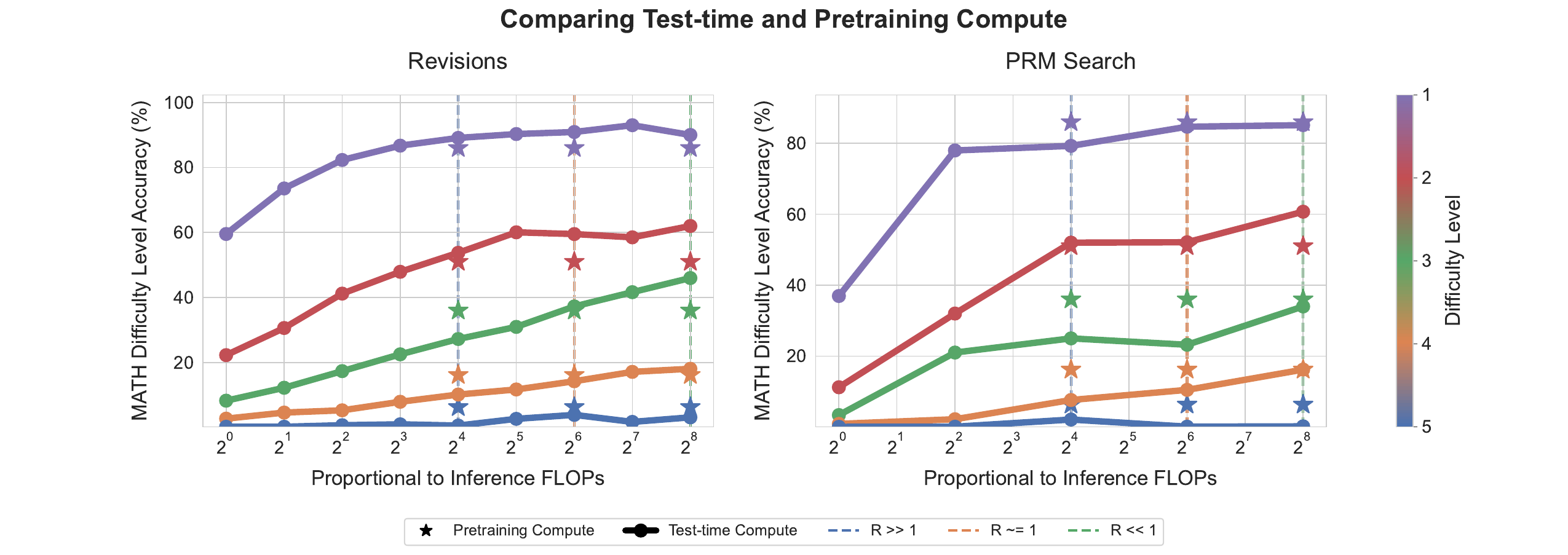}
    \vspace{-0.4cm}
    \caption{\footnotesize{\textbf{\emph{Tradeoff between pretraining and test-time compute in a FLOPs-matched evaluation.}} Each line represents the performance of scaling test-time compute with our compute-optimal policy in each oracle difficulty bin. We plot the results for revisions on the left and search on the right. The stars represent the greedy pass@1 performance of a base model pretrained with $\sim14$ times more parameters. We plot test-time compute budget on the x-axis, and place the stars at three different locations along the x-axis, each corresponding to the FLOPs equivalent point of comparison between scaling parameters and scaling test-time compute for three different inference compute loads (e.g. $R = \frac{D_{\text{inference}}}{D_{\text{pretrain}}}$). If the star is below the line, this implies that it is more effective to use test-time compute than to scale model parameters, and if the star is above the line this implies that scaling parameters is more effective. We see that on the easy questions or in settings with a lower inference load (e.g. $R << 1$), test-time compute can generally outperform scaling model parameters. However, on the harder questions or in settings with a higher inference load (e.g. $R >> 1$), pretraining is a more effective way to improve performance.}}
    \label{fig:pretrain_exchange}
    \vspace{-0.4cm}
\end{figure}

To match the FLOPs from scaling up model paramters using test-time compute with the smaller model we can multiply the smaller model's inference compute by a factor of $M + 3 \left(\frac{D_{\text{pretrain}}}{D_{\text{inference}}}\right)(M-1)$. Notably, the amount of inference compute we can utilize to match the FLOPs for the larger model depends on the ratio $\frac{D_{\text{pretrain}}}{D_{\text{inference}}}$. We refer to the inverse of this ratio as $R$ (e.g. $\frac{D_{\text{inference}}}{D_{\text{pretrain}}}$). Depending on the specific production setting or use-case, we should expect very different values of $R$. In particular, in many large scale production settings, we may expect significantly more inference tokens than pretraining tokens, in which case we would have $R >> 1$. On the other hand, in many contemporary self-improvement setups, that would use test-time compute to improve the model, we would likely generate significantly fewer inference tokens than pretraining tokens, giving $R << 1$. Therefore, since the scale of test-time compute we can apply is dependent on this ratio, we expect differing conclusions depending on the specific setting.

In Figure~\ref{fig:pretrain_exchange}, we use this approach to exchanging test-time and pretraining compute to compare our compute-optimal scaling against scaling up model parameters by a factor of $\sim14$. We conduct comparisons for 3 different values of R: 0.16 ($R << 1$), 0.79 ($R \sim 1$), and 22 ($R >> 1$), with each ratio corresponding to an inference budget. Observe that if we only expect to see very difficult questions (e.g. difficulty bins 4/5) or have a larger $D_{\text{inference}}$ (corresponding to a larger $R$ value), then it is often more effective to allocate our budget towards pretraining (e.g. the star is above the line). If instead, we expect mostly easy or intermediate difficulty questions (e.g. bins 1/2/3 and sometimes 4) or have lower inference requirements (as is the case in self-improvement pipelines), then utilizing test-time compute  is better.

\begin{AIbox}{Takeaways for exchanging pretraining and test-time compute}
Test-time and pretraining compute are not 1-to-1 ``exchangeable''. On easy and medium questions, which are within a model's capabilities, or in settings with small inference requirement, test-time compute can easily cover up for additional pretraining. However, on challenging questions which are outside a given base model's capabilities or under higher inference requirement, pretraining is  likely more effective for improving performance.
\end{AIbox}

\vspace{-0.2cm}
\section{Discussion and Future Work}
\label{sec:discussion}
\vspace{-0.2cm}

In this work, we conducted a thorough analysis of the efficacy of different techniques that aim to either improve search against a verifier or to refine an LLM's proposal distribution, for scaling test-time compute for math reasoning. In general, we found that the efficacy of a given approach heavily correlates with the difficulty of the problem from the perspective of the base LLM's capabilities. This motivated us to introduce the notion of ``compute-optimal'' scaling of test-time computation, which prescribes a adaptive, prompt-dependent strategy to improve performance under a given test-time compute budget. By applying such a compute-optimal scaling strategy, we find that can improve the efficiency of test-time compute scaling by a factor of \textbf{$2-4\times$}. When comparing benefits obtained from additional test-time compute against benefits from additional pre-training compute in a FLOPs-matched setting, we show for the first time that using test-time computation with seemingly simple methods (i.e., revisions and search) can already scale well on certain types of prompts, providing gains over spending those FLOPs in pretraining. That said, there are also limitations associated with our study that future work can aim to address. 

\textbf{Further improving test-time compute scaling.} In this work we focused on improving the test-time compute scaling of two primary mechanisms: the verifier and the proposal distribution (via revisions). While we combined verifiers with revisions in Section~\ref{sec:revisions}, we did not experiment with PRM tree-search techniques in combination with revisions. Neither did we study other techniques such as critique and revise~\citep{madaan2023selfrefine}. Future work should investigate how test-time compute scaling can be further improved by combining a variety of these approaches. Additionally, we found that across the board these schemes provided small gains on hard problems; future work should work to develop new ways of using test-time compute which can circumvent this limitation.

\textbf{Assessing question difficulty quickly.} We used a notion question difficulty as a simple sufficient statistic for approximating the compute-optimal test-time scaling strategy. While this scheme was effective, estimating our notion of difficulty requires applying a non-trivial amount of test-time compute itself. Future work should consider alternative ways of more efficiently estimating question difficulty (e.g., by pretraining or finetuning models to directly predict difficulty of a question) or dynamically switching between assessing difficulty and attempting to solve a question.

\textbf{Interleaving test-time and training-time compute.} We focused purely on test-time compute scaling in this work and the degree to which test-time compute can be traded off for additional pretraining. However, in the future, we envision that the outputs of applying additional test-time compute can be distilled back into the base LLM, enabling an iterative self-improvement loop that operates on open-ended natural language. To this end, future work should extend our findings and study how the outputs of applying test-time compute can be used to improve the base LLM itself.

\vspace{-0.2cm}
\section*{Acknowledgements}
\vspace{-0.2cm}
We thank Yi Su, Rishabh Agarwal, Yinlam Chow, Aleksandra Faust, Vincent Zhuang, George Tucker, Hao Liu, Jiayi Pan, Ethan Dyer, Behnam Neyshabur, Xavier Garcia, Yamini Bansal, Lampros Lamprou, Yuxiao Qu, and Amrith Setlur for their feedback on an earlier version of the paper and discussions. We attribute and thank Rishabh Agarwal, Vincent Zhuang, Yi Su, and Avi Singh for ideas and discussions, and experiments that concretely demonstrated the promise of pairwise sample generation for training revision models, and edit distance based sampling in \citep{anonymousrevisions}. We thank Slav Petrov for leadership support.

%% file: googledeepmind-test.bbl
\begin{thebibliography}{51}
\providecommand{\natexlab}[1]{#1}
\providecommand{\url}[1]{\texttt{#1}}
\expandafter\ifx\csname urlstyle\endcsname\relax
  \providecommand{\doi}[1]{doi: #1}\else
  \providecommand{\doi}{doi: \begingroup \urlstyle{rm}\Url}\fi

\bibitem[ano(Coming soon, 2024)]{anonymousrevisions}
Training revision models with synthetic data.
\newblock Coming soon, 2024.

\bibitem[Andrieu et~al.(2003)Andrieu, De~Freitas, Doucet, and Jordan]{andrieu2003introduction}
C.~Andrieu, N.~De~Freitas, A.~Doucet, and M.~I. Jordan.
\newblock An introduction to mcmc for machine learning.
\newblock 2003.

\bibitem[Anil et~al.(2023)Anil, Dai, Firat, Johnson, Lepikhin, Passos, Shakeri, Taropa, Bailey, Chen, Chu, Clark, Shafey, Huang, Meier-Hellstern, Mishra, Moreira, Omernick, Robinson, Ruder, Tay, Xiao, Xu, Zhang, Abrego, Ahn, Austin, Barham, Botha, Bradbury, Brahma, Brooks, Catasta, Cheng, Cherry, Choquette-Choo, Chowdhery, Crepy, Dave, Dehghani, Dev, Devlin, Díaz, Du, Dyer, Feinberg, Feng, Fienber, Freitag, Garcia, Gehrmann, Gonzalez, Gur-Ari, Hand, Hashemi, Hou, Howland, Hu, Hui, Hurwitz, Isard, Ittycheriah, Jagielski, Jia, Kenealy, Krikun, Kudugunta, Lan, Lee, Lee, Li, Li, Li, Li, Li, Lim, Lin, Liu, Liu, Maggioni, Mahendru, Maynez, Misra, Moussalem, Nado, Nham, Ni, Nystrom, Parrish, Pellat, Polacek, Polozov, Pope, Qiao, Reif, Richter, Riley, Ros, Roy, Saeta, Samuel, Shelby, Slone, Smilkov, So, Sohn, Tokumine, Valter, Vasudevan, Vodrahalli, Wang, Wang, Wang, Wang, Wieting, Wu, Xu, Xu, Xue, Yin, Yu, Zhang, Zheng, Zheng, Zhou, Zhou, Petrov, and Wu]{anil2023palm}
R.~Anil, A.~M. Dai, O.~Firat, M.~Johnson, D.~Lepikhin, A.~Passos, S.~Shakeri, E.~Taropa, P.~Bailey, Z.~Chen, E.~Chu, J.~H. Clark, L.~E. Shafey, Y.~Huang, K.~Meier-Hellstern, G.~Mishra, E.~Moreira, M.~Omernick, K.~Robinson, S.~Ruder, Y.~Tay, K.~Xiao, Y.~Xu, Y.~Zhang, G.~H. Abrego, J.~Ahn, J.~Austin, P.~Barham, J.~Botha, J.~Bradbury, S.~Brahma, K.~Brooks, M.~Catasta, Y.~Cheng, C.~Cherry, C.~A. Choquette-Choo, A.~Chowdhery, C.~Crepy, S.~Dave, M.~Dehghani, S.~Dev, J.~Devlin, M.~Díaz, N.~Du, E.~Dyer, V.~Feinberg, F.~Feng, V.~Fienber, M.~Freitag, X.~Garcia, S.~Gehrmann, L.~Gonzalez, G.~Gur-Ari, S.~Hand, H.~Hashemi, L.~Hou, J.~Howland, A.~Hu, J.~Hui, J.~Hurwitz, M.~Isard, A.~Ittycheriah, M.~Jagielski, W.~Jia, K.~Kenealy, M.~Krikun, S.~Kudugunta, C.~Lan, K.~Lee, B.~Lee, E.~Li, M.~Li, W.~Li, Y.~Li, J.~Li, H.~Lim, H.~Lin, Z.~Liu, F.~Liu, M.~Maggioni, A.~Mahendru, J.~Maynez, V.~Misra, M.~Moussalem, Z.~Nado, J.~Nham, E.~Ni, A.~Nystrom, A.~Parrish, M.~Pellat, M.~Polacek, A.~Polozov, R.~Pope, S.~Qiao, E.~Reif, B.~Richter,
  P.~Riley, A.~C. Ros, A.~Roy, B.~Saeta, R.~Samuel, R.~Shelby, A.~Slone, D.~Smilkov, D.~R. So, D.~Sohn, S.~Tokumine, D.~Valter, V.~Vasudevan, K.~Vodrahalli, X.~Wang, P.~Wang, Z.~Wang, T.~Wang, J.~Wieting, Y.~Wu, K.~Xu, Y.~Xu, L.~Xue, P.~Yin, J.~Yu, Q.~Zhang, S.~Zheng, C.~Zheng, W.~Zhou, D.~Zhou, S.~Petrov, and Y.~Wu.
\newblock Palm 2 technical report, 2023.

\bibitem[Bai et~al.(2022)Bai, Kadavath, Kundu, Askell, Kernion, Jones, Chen, Goldie, Mirhoseini, McKinnon, Chen, Olsson, Olah, Hernandez, Drain, Ganguli, Li, Tran-Johnson, Perez, Kerr, Mueller, Ladish, Landau, Ndousse, Lukosuite, Lovitt, Sellitto, Elhage, Schiefer, Mercado, DasSarma, Lasenby, Larson, Ringer, Johnston, Kravec, Showk, Fort, Lanham, Telleen-Lawton, Conerly, Henighan, Hume, Bowman, Hatfield-Dodds, Mann, Amodei, Joseph, McCandlish, Brown, and Kaplan]{bai2022constitutional}
Y.~Bai, S.~Kadavath, S.~Kundu, A.~Askell, J.~Kernion, A.~Jones, A.~Chen, A.~Goldie, A.~Mirhoseini, C.~McKinnon, C.~Chen, C.~Olsson, C.~Olah, D.~Hernandez, D.~Drain, D.~Ganguli, D.~Li, E.~Tran-Johnson, E.~Perez, J.~Kerr, J.~Mueller, J.~Ladish, J.~Landau, K.~Ndousse, K.~Lukosuite, L.~Lovitt, M.~Sellitto, N.~Elhage, N.~Schiefer, N.~Mercado, N.~DasSarma, R.~Lasenby, R.~Larson, S.~Ringer, S.~Johnston, S.~Kravec, S.~E. Showk, S.~Fort, T.~Lanham, T.~Telleen-Lawton, T.~Conerly, T.~Henighan, T.~Hume, S.~R. Bowman, Z.~Hatfield-Dodds, B.~Mann, D.~Amodei, N.~Joseph, S.~McCandlish, T.~Brown, and J.~Kaplan.
\newblock Constitutional ai: Harmlessness from ai feedback, 2022.

\bibitem[Blakeney et~al.(2024)Blakeney, Paul, Larsen, Owen, and Frankle]{blakeney2024doesdatasparkjoy}
C.~Blakeney, M.~Paul, B.~W. Larsen, S.~Owen, and J.~Frankle.
\newblock Does your data spark joy? performance gains from domain upsampling at the end of training, 2024.
\newblock URL \url{https://arxiv.org/abs/2406.03476}.

\bibitem[Chen et~al.(2024)Chen, Liao, Li, and Fan]{chen2024alphamath}
G.~Chen, M.~Liao, C.~Li, and K.~Fan.
\newblock Alphamath almost zero: process supervision without process, 2024.

\bibitem[Cobbe et~al.(2021)Cobbe, Kosaraju, Bavarian, Chen, Jun, Kaiser, Plappert, Tworek, Hilton, Nakano, Hesse, and Schulman]{cobbe2021training}
K.~Cobbe, V.~Kosaraju, M.~Bavarian, M.~Chen, H.~Jun, L.~Kaiser, M.~Plappert, J.~Tworek, J.~Hilton, R.~Nakano, C.~Hesse, and J.~Schulman.
\newblock Training verifiers to solve math word problems, 2021.

\bibitem[Du et~al.(2023)Du, Li, Torralba, Tenenbaum, and Mordatch]{du2023improving}
Y.~Du, S.~Li, A.~Torralba, J.~B. Tenenbaum, and I.~Mordatch.
\newblock Improving factuality and reasoning in language models through multiagent debate, 2023.

\bibitem[Evans(1984)]{evans1984heuristic}
J.~S. B.~T. Evans.
\newblock Heuristic and analytic processes in reasoning.
\newblock \emph{British Journal of Psychology}, 75(4):\penalty0 451--468, 1984.

\bibitem[Feng et~al.(2024)Feng, Wan, Wen, McAleer, Wen, Zhang, and Wang]{feng2024alphazerolike}
X.~Feng, Z.~Wan, M.~Wen, S.~M. McAleer, Y.~Wen, W.~Zhang, and J.~Wang.
\newblock Alphazero-like tree-search can guide large language model decoding and training, 2024.

\bibitem[Gao et~al.(2023)Gao, Madaan, Zhou, Alon, Liu, Yang, Callan, and Neubig]{gao2023palprogramaidedlanguagemodels}
L.~Gao, A.~Madaan, S.~Zhou, U.~Alon, P.~Liu, Y.~Yang, J.~Callan, and G.~Neubig.
\newblock Pal: Program-aided language models, 2023.
\newblock URL \url{https://arxiv.org/abs/2211.10435}.

\bibitem[Goyal et~al.(2024)Goyal, Ji, Rawat, Menon, Kumar, and Nagarajan]{goyal2024thinkspeaktraininglanguage}
S.~Goyal, Z.~Ji, A.~S. Rawat, A.~K. Menon, S.~Kumar, and V.~Nagarajan.
\newblock Think before you speak: Training language models with pause tokens, 2024.
\newblock URL \url{https://arxiv.org/abs/2310.02226}.

\bibitem[Hendrycks et~al.(2021)Hendrycks, Burns, Kadavath, Arora, Basart, Tang, Song, and Steinhardt]{hendrycks2021measuring}
D.~Hendrycks, C.~Burns, S.~Kadavath, A.~Arora, S.~Basart, E.~Tang, D.~Song, and J.~Steinhardt.
\newblock Measuring mathematical problem solving with the math dataset, 2021.

\bibitem[Hoffmann et~al.(2022)Hoffmann, Borgeaud, Mensch, Buchatskaya, Cai, Rutherford, de~Las~Casas, Hendricks, Welbl, Clark, Hennigan, Noland, Millican, van~den Driessche, Damoc, Guy, Osindero, Simonyan, Elsen, Rae, Vinyals, and Sifre]{hoffmann2022training}
J.~Hoffmann, S.~Borgeaud, A.~Mensch, E.~Buchatskaya, T.~Cai, E.~Rutherford, D.~de~Las~Casas, L.~A. Hendricks, J.~Welbl, A.~Clark, T.~Hennigan, E.~Noland, K.~Millican, G.~van~den Driessche, B.~Damoc, A.~Guy, S.~Osindero, K.~Simonyan, E.~Elsen, J.~W. Rae, O.~Vinyals, and L.~Sifre.
\newblock Training compute-optimal large language models, 2022.

\bibitem[Huang et~al.(2023)Huang, Chen, Mishra, Zheng, Yu, Song, and Zhou]{huang2023large}
J.~Huang, X.~Chen, S.~Mishra, H.~S. Zheng, A.~W. Yu, X.~Song, and D.~Zhou.
\newblock Large language models cannot self-correct reasoning yet, 2023.

\bibitem[Jones(2021)]{jones2021scalingscalinglawsboard}
A.~L. Jones.
\newblock Scaling scaling laws with board games, 2021.
\newblock URL \url{https://arxiv.org/abs/2104.03113}.

\bibitem[Kahneman(2003)]{kahneman2003maps}
D.~Kahneman.
\newblock Maps of bounded rationality: Psychology for behavioral economics.
\newblock \emph{The American Economic Review}, 93(5):\penalty0 1449--1475, 2003.

\bibitem[Kahneman(2013)]{kahneman2013thinking}
D.~Kahneman.
\newblock \emph{Thinking, fast and slow}.
\newblock Farrar, Straus and Giroux, New York, first paperback edition edition, 2013.

\bibitem[Kocsis and Szepesv{'a}ri(2006)]{kocsis2006bandit}
L.~Kocsis and C.~Szepesv{'a}ri.
\newblock Bandit based monte-carlo planning.
\newblock In \emph{European conference on machine learning}, pages 282--293. Springer, 2006.

\bibitem[Lewkowycz et~al.(2022)Lewkowycz, Andreassen, Dohan, Dyer, Michalewski, Ramasesh, Slone, Anil, Schlag, Gutman-Solo, Wu, Neyshabur, Gur-Ari, and Misra]{lewkowycz2022solving}
A.~Lewkowycz, A.~Andreassen, D.~Dohan, E.~Dyer, H.~Michalewski, V.~Ramasesh, A.~Slone, C.~Anil, I.~Schlag, T.~Gutman-Solo, Y.~Wu, B.~Neyshabur, G.~Gur-Ari, and V.~Misra.
\newblock Solving quantitative reasoning problems with language models, 2022.

\bibitem[Li et~al.(2023)Li, Lin, Zhang, Fu, Chen, Lou, and Chen]{li2023making}
Y.~Li, Z.~Lin, S.~Zhang, Q.~Fu, B.~Chen, J.-G. Lou, and W.~Chen.
\newblock Making large language models better reasoners with step-aware verifier, 2023.

\bibitem[Lightman et~al.(2023)Lightman, Kosaraju, Burda, Edwards, Baker, Lee, Leike, Schulman, Sutskever, and Cobbe]{lightman2023lets}
H.~Lightman, V.~Kosaraju, Y.~Burda, H.~Edwards, B.~Baker, T.~Lee, J.~Leike, J.~Schulman, I.~Sutskever, and K.~Cobbe.
\newblock Let's verify step by step, 2023.

\bibitem[Madaan et~al.(2023)Madaan, Tandon, Gupta, Hallinan, Gao, Wiegreffe, Alon, Dziri, Prabhumoye, Yang, Gupta, Majumder, Hermann, Welleck, Yazdanbakhsh, and Clark]{madaan2023selfrefine}
A.~Madaan, N.~Tandon, P.~Gupta, S.~Hallinan, L.~Gao, S.~Wiegreffe, U.~Alon, N.~Dziri, S.~Prabhumoye, Y.~Yang, S.~Gupta, B.~P. Majumder, K.~Hermann, S.~Welleck, A.~Yazdanbakhsh, and P.~Clark.
\newblock Self-refine: Iterative refinement with self-feedback, 2023.

\bibitem[McAleese et~al.(2024)McAleese, Pokorny, Cerón~Uribe, Nitishinskaya, Trębacz, and Leike]{llmcriticscatchbugs}
N.~McAleese, R.~Pokorny, J.~F. Cerón~Uribe, E.~Nitishinskaya, M.~Trębacz, and J.~Leike.
\newblock Llm critics help catch llm bugs.
\newblock \emph{OpenAI}, 2024.

\bibitem[OpenAI(2024)]{openai2024gpt4}
OpenAI.
\newblock Gpt-4 technical report, 2024.

\bibitem[Qin et~al.(2023)Qin, Liang, Ye, Zhu, Yan, Lu, Lin, Cong, Tang, Qian, Zhao, Hong, Tian, Xie, Zhou, Gerstein, Li, Liu, and Sun]{qin2023toolllmfacilitatinglargelanguage}
Y.~Qin, S.~Liang, Y.~Ye, K.~Zhu, L.~Yan, Y.~Lu, Y.~Lin, X.~Cong, X.~Tang, B.~Qian, S.~Zhao, L.~Hong, R.~Tian, R.~Xie, J.~Zhou, M.~Gerstein, D.~Li, Z.~Liu, and M.~Sun.
\newblock Toolllm: Facilitating large language models to master 16000+ real-world apis, 2023.
\newblock URL \url{https://arxiv.org/abs/2307.16789}.

\bibitem[Qu et~al.(2024{\natexlab{a}})Qu, Dai, Wei, Cai, Wang, Yin, Xu, and Wen]{qu2024toollearninglargelanguage}
C.~Qu, S.~Dai, X.~Wei, H.~Cai, S.~Wang, D.~Yin, J.~Xu, and J.-R. Wen.
\newblock Tool learning with large language models: A survey, 2024{\natexlab{a}}.
\newblock URL \url{https://arxiv.org/abs/2405.17935}.

\bibitem[Qu et~al.(2024{\natexlab{b}})Qu, Zhang, Garg, and Kumar]{qu2024recursive}
Y.~Qu, T.~Zhang, N.~Garg, and A.~Kumar.
\newblock Recursive introspection: Teaching foundation models how to self-improve.
\newblock 2024{\natexlab{b}}.

\bibitem[Sardana and Frankle(2023)]{sardana2023chinchillaoptimal}
N.~Sardana and J.~Frankle.
\newblock Beyond chinchilla-optimal: Accounting for inference in language model scaling laws, 2023.

\bibitem[Saunders et~al.(2022)Saunders, Yeh, Wu, Bills, Ouyang, Ward, and Leike]{saunders2022selfcritiquing}
W.~Saunders, C.~Yeh, J.~Wu, S.~Bills, L.~Ouyang, J.~Ward, and J.~Leike.
\newblock Self-critiquing models for assisting human evaluators, 2022.

\bibitem[Setlur et~al.(2024)Setlur, Garg, Geng, Garg, Smith, and Kumar]{setlur2024rl}
A.~Setlur, S.~Garg, X.~Geng, N.~Garg, V.~Smith, and A.~Kumar.
\newblock Rl on incorrect synthetic data scales the efficiency of llm math reasoning by eight-fold.
\newblock \emph{arXiv preprint arXiv:2406.14532}, 2024.

\bibitem[Shao et~al.(2024)Shao, Wang, Zhu, Xu, Song, Bi, Zhang, Zhang, Li, Wu, and Guo]{shao2024deepseekmath}
Z.~Shao, P.~Wang, Q.~Zhu, R.~Xu, J.~Song, X.~Bi, H.~Zhang, M.~Zhang, Y.~K. Li, Y.~Wu, and D.~Guo.
\newblock Deepseekmath: Pushing the limits of mathematical reasoning in open language models, 2024.

\bibitem[Sharma et~al.(2024)Sharma, Keh, Mitchell, Finn, Arora, and Kollar]{sharma2024criticalevaluationaifeedback}
A.~Sharma, S.~Keh, E.~Mitchell, C.~Finn, K.~Arora, and T.~Kollar.
\newblock A critical evaluation of ai feedback for aligning large language models, 2024.
\newblock URL \url{https://arxiv.org/abs/2402.12366}.

\bibitem[Shinn et~al.(2023)Shinn, Cassano, Berman, Gopinath, Narasimhan, and Yao]{shinn2023reflexion}
N.~Shinn, F.~Cassano, E.~Berman, A.~Gopinath, K.~Narasimhan, and S.~Yao.
\newblock Reflexion: Language agents with verbal reinforcement learning, 2023.

\bibitem[Singh et~al.(2024)Singh, Co-Reyes, Agarwal, Anand, Patil, Garcia, Liu, Harrison, Lee, Xu, Parisi, Kumar, Alemi, Rizkowsky, Nova, Adlam, Bohnet, Elsayed, Sedghi, Mordatch, Simpson, Gur, Snoek, Pennington, Hron, Kenealy, Swersky, Mahajan, Culp, Xiao, Bileschi, Constant, Novak, Liu, Warkentin, Qian, Bansal, Dyer, Neyshabur, Sohl-Dickstein, and Fiedel]{singh2024human}
A.~Singh, J.~D. Co-Reyes, R.~Agarwal, A.~Anand, P.~Patil, X.~Garcia, P.~J. Liu, J.~Harrison, J.~Lee, K.~Xu, A.~Parisi, A.~Kumar, A.~Alemi, A.~Rizkowsky, A.~Nova, B.~Adlam, B.~Bohnet, G.~Elsayed, H.~Sedghi, I.~Mordatch, I.~Simpson, I.~Gur, J.~Snoek, J.~Pennington, J.~Hron, K.~Kenealy, K.~Swersky, K.~Mahajan, L.~Culp, L.~Xiao, M.~L. Bileschi, N.~Constant, R.~Novak, R.~Liu, T.~Warkentin, Y.~Qian, Y.~Bansal, E.~Dyer, B.~Neyshabur, J.~Sohl-Dickstein, and N.~Fiedel.
\newblock Beyond human data: Scaling self-training for problem-solving with language models, 2024.

\bibitem[Snell et~al.(2024)Snell, Wallace, Klein, and Levine]{predictemergence}
C.~Snell, E.~Wallace, D.~Klein, and S.~Levine.
\newblock Predicting emergent capabilities by finetuning.
\newblock \emph{Conference on Language Modeling 2024}, 2024.

\bibitem[Stechly et~al.(2023)Stechly, Marquez, and Kambhampati]{stechly2023gpt4}
K.~Stechly, M.~Marquez, and S.~Kambhampati.
\newblock Gpt-4 doesn't know it's wrong: An analysis of iterative prompting for reasoning problems, 2023.

\bibitem[Sutton and Barto(2018)]{suttonrlbook}
R.~S. Sutton and A.~G. Barto.
\newblock \emph{Reinforcement learning: An introduction}.
\newblock Second edition, 2018.

\bibitem[Team(2024)]{geminiteam2024gemini}
G.~Team.
\newblock Gemini 1.5: Unlocking multimodal understanding across millions of tokens of context, 2024.

\bibitem[Tian et~al.(2024)Tian, Peng, Song, Jin, Yu, Mi, and Yu]{tian2024selfimprovement}
Y.~Tian, B.~Peng, L.~Song, L.~Jin, D.~Yu, H.~Mi, and D.~Yu.
\newblock Toward self-improvement of llms via imagination, searching, and criticizing, 2024.

\bibitem[Touvron et~al.(2023)Touvron, Martin, Stone, Albert, Almahairi, Babaei, Bashlykov, Batra, Bhargava, Bhosale, Bikel, Blecher, Ferrer, Chen, Cucurull, Esiobu, Fernandes, Fu, Fu, Fuller, Gao, Goswami, Goyal, Hartshorn, Hosseini, Hou, Inan, Kardas, Kerkez, Khabsa, Kloumann, Korenev, Koura, Lachaux, Lavril, Lee, Liskovich, Lu, Mao, Martinet, Mihaylov, Mishra, Molybog, Nie, Poulton, Reizenstein, Rungta, Saladi, Schelten, Silva, Smith, Subramanian, Tan, Tang, Taylor, Williams, Kuan, Xu, Yan, Zarov, Zhang, Fan, Kambadur, Narang, Rodriguez, Stojnic, Edunov, and Scialom]{touvron2023llama2openfoundation}
H.~Touvron, L.~Martin, K.~Stone, P.~Albert, A.~Almahairi, Y.~Babaei, N.~Bashlykov, S.~Batra, P.~Bhargava, S.~Bhosale, D.~Bikel, L.~Blecher, C.~C. Ferrer, M.~Chen, G.~Cucurull, D.~Esiobu, J.~Fernandes, J.~Fu, W.~Fu, B.~Fuller, C.~Gao, V.~Goswami, N.~Goyal, A.~Hartshorn, S.~Hosseini, R.~Hou, H.~Inan, M.~Kardas, V.~Kerkez, M.~Khabsa, I.~Kloumann, A.~Korenev, P.~S. Koura, M.-A. Lachaux, T.~Lavril, J.~Lee, D.~Liskovich, Y.~Lu, Y.~Mao, X.~Martinet, T.~Mihaylov, P.~Mishra, I.~Molybog, Y.~Nie, A.~Poulton, J.~Reizenstein, R.~Rungta, K.~Saladi, A.~Schelten, R.~Silva, E.~M. Smith, R.~Subramanian, X.~E. Tan, B.~Tang, R.~Taylor, A.~Williams, J.~X. Kuan, P.~Xu, Z.~Yan, I.~Zarov, Y.~Zhang, A.~Fan, M.~Kambadur, S.~Narang, A.~Rodriguez, R.~Stojnic, S.~Edunov, and T.~Scialom.
\newblock Llama 2: Open foundation and fine-tuned chat models, 2023.
\newblock URL \url{https://arxiv.org/abs/2307.09288}.

\bibitem[Uesato et~al.(2022)Uesato, Kushman, Kumar, Song, Siegel, Wang, Creswell, Irving, and Higgins]{uesato2022solving}
J.~Uesato, N.~Kushman, R.~Kumar, F.~Song, N.~Siegel, L.~Wang, A.~Creswell, G.~Irving, and I.~Higgins.
\newblock Solving math word problems with process- and outcome-based feedback, 2022.

\bibitem[Valmeekam et~al.(2023)Valmeekam, Marquez, and Kambhampati]{valmeekam2023large}
K.~Valmeekam, M.~Marquez, and S.~Kambhampati.
\newblock Can large language models really improve by self-critiquing their own plans?, 2023.

\bibitem[Villalobos and Atkinson(2023)]{epoch2023tradingoffcomputeintrainingandinference}
P.~Villalobos and D.~Atkinson.
\newblock Trading off compute in training and inference, 2023.
\newblock URL \url{https://epochai.org/blog/trading-off-compute-in-training-and-inference}.
\newblock Accessed: 2024-07-03.

\bibitem[Wang et~al.(2023)Wang, Li, Shao, Xu, Dai, Li, Chen, Wu, and Sui]{wang2023mathshepherd}
P.~Wang, L.~Li, Z.~Shao, R.~X. Xu, D.~Dai, Y.~Li, D.~Chen, Y.~Wu, and Z.~Sui.
\newblock Math-shepherd: Verify and reinforce llms step-by-step without human annotations, 2023.

\bibitem[Wang et~al.(2024)Wang, Zelikman, Poesia, Pu, Haber, and Goodman]{wang2024hypothesissearchinductivereasoning}
R.~Wang, E.~Zelikman, G.~Poesia, Y.~Pu, N.~Haber, and N.~D. Goodman.
\newblock Hypothesis search: Inductive reasoning with language models, 2024.
\newblock URL \url{https://arxiv.org/abs/2309.05660}.

\bibitem[Wei et~al.(2023)Wei, Wang, Schuurmans, Bosma, Ichter, Xia, Chi, Le, and Zhou]{wei2023chainofthought}
J.~Wei, X.~Wang, D.~Schuurmans, M.~Bosma, B.~Ichter, F.~Xia, E.~Chi, Q.~Le, and D.~Zhou.
\newblock Chain-of-thought prompting elicits reasoning in large language models, 2023.

\bibitem[Yao et~al.(2023)Yao, Yu, Zhao, Shafran, Griffiths, Cao, and Narasimhan]{yao2023tree}
S.~Yao, D.~Yu, J.~Zhao, I.~Shafran, T.~L. Griffiths, Y.~Cao, and K.~Narasimhan.
\newblock Tree of thoughts: Deliberate problem solving with large language models, 2023.

\bibitem[Yuan et~al.(2023)Yuan, Yuan, Li, Dong, Lu, Tan, Zhou, and Zhou]{yuan2023scaling}
Z.~Yuan, H.~Yuan, C.~Li, G.~Dong, K.~Lu, C.~Tan, C.~Zhou, and J.~Zhou.
\newblock Scaling relationship on learning mathematical reasoning with large language models, 2023.

\bibitem[Zelikman et~al.(2022)Zelikman, Wu, Mu, and Goodman]{zelikman2022star}
E.~Zelikman, Y.~Wu, J.~Mu, and N.~D. Goodman.
\newblock Star: Bootstrapping reasoning with reasoning, 2022.

\bibitem[Zelikman et~al.(2024)Zelikman, Harik, Shao, Jayasiri, Haber, and Goodman]{zelikman2024quietstarlanguagemodelsteach}
E.~Zelikman, G.~Harik, Y.~Shao, V.~Jayasiri, N.~Haber, and N.~D. Goodman.
\newblock Quiet-star: Language models can teach themselves to think before speaking, 2024.
\newblock URL \url{https://arxiv.org/abs/2403.09629}.

\end{thebibliography}
